\documentclass{article}

\PassOptionsToPackage{numbers, compress}{natbib}
\usepackage[final]{neurips_2023}

\usepackage[utf8]{inputenc} % allow utf-8 input
\usepackage[T1]{fontenc}    % use 8-bit T1 fonts
\usepackage{hyperref}       % hyperlinks
\usepackage{url}            % simple URL typesetting
\usepackage{booktabs}       % professional-quality tables
\usepackage{amsfonts}       % blackboard math symbols
\usepackage{nicefrac}       % compact symbols for 1/2, etc.
\usepackage{microtype}      % microtypography
\usepackage{xcolor}         % colors
\usepackage{multirow}
\usepackage{siunitx}
\usepackage{tikz}
\usepackage{standalone}
\usepackage{import}
\usepackage{caption} 
\usepackage{subcaption}
\usepackage{outlines}
\usepackage{hhline}
\usepackage{amsmath}
\usepackage{mathtools}
\usepackage{pgfplots}
\pgfplotsset{compat=1.18}
\usepackage{tabularx}
\usepackage{makecell}
\usepackage{siunitx}
\usepackage{changepage}
\usepackage[ruled,linesnumbered]{algorithm2e}
\usepackage{afterpage}
\usepackage[utf8]{inputenc}
\usepackage{algorithmic}
\usepackage{setspace}
\usepackage{placeins}

\newcommand{\R}{\mathbb R}
\DeclareMathOperator{\logsumexp}{\text{LogSumExp}}
\newcommand{\xs}{x_\text{synth}}
\newcommand{\ys}{y_\text{synth}}
\newcommand{\xtr}{x_\text{train}}
\newcommand{\ytr}{y_\text{train}}
\newcommand{\xte}{x_\text{test}}

\newcommand{\ytep}{\hat{y}_\text{test}}

\title{TabPFGen -- Tabular Data Generation with TabPFN}

\author{%
  Junwei Ma, Apoorv Dankar, George Stein, Guangwei Yu, Anthony Caterini \\
  Layer 6 AI \\
  Toronto, Canada \\
  \texttt{\{jeremy, apoorv, george, guang, anthony\}@layer6.ai} \\
}

\begin{document}

\maketitle

\begin{abstract}
  Advances in deep generative modelling have not translated well to tabular data. We argue that this is caused by a mismatch in structure between popular generative models, and \emph{discriminative} models of tabular data. We thus devise a technique to turn TabPFN -- a highly performant transformer initially designed for in-context discriminative tabular tasks -- into an energy-based generative model, which we dub \emph{TabPFGen}. This novel framework leverages the pre-trained TabPFN as part of the energy function and does not require any additional training or hyperparameter tuning, thus inheriting TabPFN's in-context learning capability. We can sample from TabPFGen analogously to other energy-based models. We demonstrate strong results on standard generative modelling tasks, including data augmentation, class-balancing, and imputation, unlocking a new frontier of tabular data generation.
  
\end{abstract}

% Our approach relies on noisy optimization -- in feature space, not parameter space -- of the classification objective, which can be seen as sampling a particular energy-based model.
%sampling, data augmentation, and class-balancing.

% However, advances in deep generative modelling on data such as images \citep{brock2018large, vahdat2020nvae, rombach2022high} and text \citep{devlin2018bert, ma2019flowseq} have not translated to the tabular setting \citep{manousakas2023usefulness}.

% It is therefore worth considering if we can leverage this model for \emph{generative} tasks, such as data augmentation and class balancing.
% We answer this in the affirmative by introducing \emph{TabPFGen} as a powerful tool to generate in-distribution data using a pre-trained TabPFN classifier.
% TabPFGen is a novel energy-based model, which defines a class-conditional generative energy in terms of a cross-entropy objective used within the frozen TabPFN; we then generate samples from this model using a similar method to stochastic gradient Langevin dynamics \citep{welling2011bayesian}.
% We note that we inherit the in-context learning capabilities of TabPFN: we do not perform any training of TabPFGen, only sampling.

% These discriminative models are typically tree-based -- thus not differentiable -- which makes it challenging to use them in gradient-based methods.

\section{Introduction}

Tabular data is pervasive and important across various domains \citep{benjelloun2020google, ulmer2020trust, clements2020sequential, tang2020customer, urban2021deep}, yet the application of deep generative modelling -- successful in modalities like images \citep{brock2018large, vahdat2020nvae, rombach2022high} and text \citep{devlin2018bert, ma2019flowseq} -- has lagged behind in the tabular setting \citep{manousakas2023usefulness}.
Previous works \citep{tu2007learning, grathwohl2020your, nock2022generative} have argued that attempts in this direction have not used high performing \emph{discriminative} models of tabular data effectively.
These discriminative models can assume various forms, encompassing tree-based and deep learning models.
The non-differentiability of tree-based models makes it challenging to integrate them into gradient-based methods.
Conversely, conventional deep learning-based models demand substantial time and effort in terms of training and hyperparameter tuning, rendering them hard to use across diverse datasets \citep{shwartz2022tabular}.
An exception to this trend is TabPFN \citep{hollmann2023tabpfn}, a transformer-based model designed for tabular data, which has demonstrated remarkable in-context learning capability for discriminative tasks. %on tabular data.
It is thus worth considering if TabPFN can be leveraged for \emph{generative} tasks.

We answer this in the affirmative by introducing \emph{TabPFGen}, a novel energy-based model that harnesses the power of TabPFN for tabular tasks in generative modelling. %data synthesis tasks including data augmentation, class balancing, and imputation.
TabPFGen defines a class-conditional energy within the pre-trained TabPFN, and employs the workhorse stochastic gradient Langevin dynamics algorithm \citep{welling2011bayesian} for sample generation. Notably, TabPFGen inherits TabPFN's in-context learning capabilities, requiring no additional training or hyperparameter tuning. We conduct experiments on 18 well-established datasets from OpenML-CC18 \citep{bischl2021openml}. Our results show a substantial improvement in downstream model performance through the utilization of TabPFGen for data augmentation, surpassing competitive baselines. Moreover, TabPFGen also proves valuable for class balancing and imputation by producing samples that closely align with the training data distribution, showcasing its exciting potential for tabular data generation in practice.

\section{Background \& Related Work}
\label{gen_inst}

\textbf{TabPFN}, introduced by \citet{hollmann2023tabpfn}, is a transformer-based architecture designed for in-context learning of discriminative tabular data tasks. It is trained using a prior-fitting procedure \citep{muller2022transformers} that exposes the network to a massive number of possible inductive biases which may be observed in the tabular setting. 
After training, the learned TabPFN model accepts training samples ($\xtr$, $\ytr$) and test features $\xte$, and yields predictions $\ytep$ for the entire test set in a single forward pass.
% After training, the learned TabPFN model can be used to generate a posterior predictive distribution over test labels $\yte$ given test features $\xte$, alongside training labels $\ytr$ and features $\xtr$. 
In our work, we probabilistically invert these inductive biases in order to generate synthetic data $\xs$ conditioned on synthetic labels $\ys$.

The field of \textbf{generative modeling for tabular data} has witnessed significant advancements. Initially, GAN-based approaches \citep{jordon2018pate, xu2019modeling, engelmann2021conditional, nock2022generative} dominated, followed by diffusion models \citep{zheng2022diffusion, kotelnikov2023tabddpm} and large language models (LLMs) \citep{borisov2022language, solatorio2023realtabformer}.
However, surprisingly, the simplest interpolation methods such as SMOTE \citep{chawla2002smote, more2016survey} remain competitive \citep{pmlr-v137-camino20a}. %still prove to be very competitive \citep{pmlr-v137-camino20a, manousakas2023usefulness}.
We conjecture that the aforementioned generative techniques may not have adequately captured the inductive biases of successful \emph{discriminative} approaches, undermining their effectiveness.
Meanwhile, \textbf{generative modelling using discriminators} has expanded mainly outside of the tabular domain. % over the years.
Early work by \citet{tu2007learning} showed promising results on computer vision tasks, and recent investigations have further demonstrated efficacy in image synthesis \citep{santurkar2019image, li2022use}. \citet{nock2022generative} also shed light on this strategy for tabular data synthesis using decision trees.%, inspiring our approach.

\textbf{Energy-based models (EBMs)} have also gained significant traction across machine learning domains. For example, \citet{florence2022implicit} utilized EBMs in the context of robot behavioral cloning, while \citet{liu2020energy} applied an energy score for out-of-distribution detection.
It is worth noting that energy scores inherently lack class information; 
our contribution builds upon the existing work by introducing a class-conditional energy for sample generation. %conditioned on the class. 
\citet{grathwohl2020your} highlighted that any classifier can be treated as an EBM, a concept that has been applied to various image generation works \citep{xie2016theory, gao2017learning, du2019implicit, yin2021see}.
Notably, our work stands out by generating tabular data while leveraging a pre-trained model without any additional training or hyperparameter tuning.

\section{Method}
\label{headings}

% General Problems:
% \begin{itemize}
%     \item need to add equations for this
%     \item should call this: train-gen or train-synth? synth is more aligned with past works 
% \end{itemize}

% \begin{figure}[h]
%      \centering
%      \begin{subfigure}[b]{0.45\textwidth}
%          \centering
%          \includegraphics[width=\textwidth]{train_gen.png}
%          \caption{Train-Synth Setup}
%          \label{fig:train-synth}
%      \end{subfigure}
%      % \hfill
%      \begin{subfigure}[b]{0.45\textwidth}
%          \centering
%          \includegraphics[width=\textwidth]{gen_train.png}
%          \caption{Synth-Train Setup}
%          \label{fig:synth-train}
%      \end{subfigure}
%      \caption{TabPFGen Overview}
%      \label{fig:tabpfgen}
% \end{figure}

% \subsection{Motivation}

% TabPFGen\textsubscript{core} Overview. While keeping TabPFN frozen, we backpropagate from the class-conditional energy to $\xs$, in order to generate gradients for SGLD and thus sample from $p(\xs \mid \ys) \propto \exp(-E(\xs \mid \ys))$. CE represents cross entropy. Blue and red arrows represent attention.

% \vspace{-7pt}
\begin{figure}[h]
     \centering
     \includegraphics[width=\textwidth]{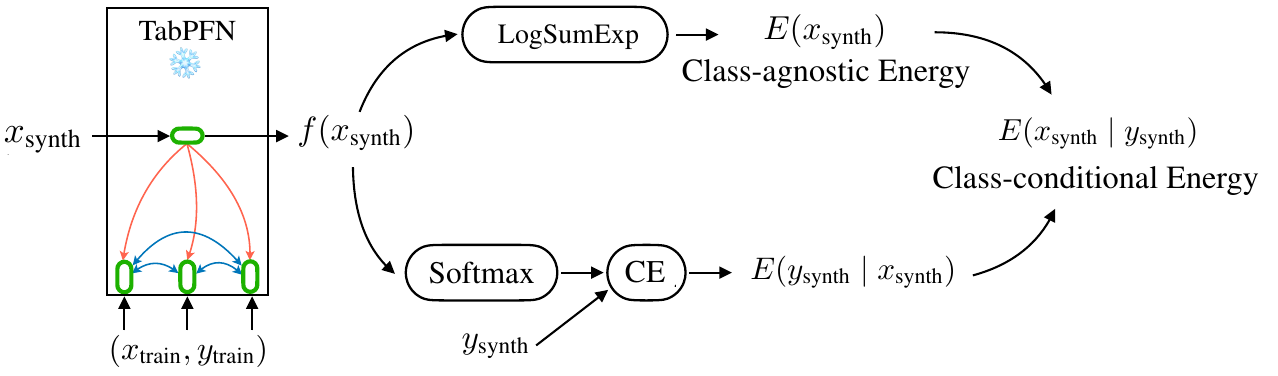}
     \caption{TabPFGen overview. We backpropagate from the class-conditional energy to $\xs$ for gradient generation. CE denotes cross entropy; blue and red arrows represent attention.}
     \label{fig:tabpfgen}
\end{figure}
% \vspace{-5pt}

We leverage the strong in-context discriminative performance of TabPFN to devise a class-conditional generative model.
In particular, given a synthetic label $\ys \in \{1,\ldots, K\}$, we seek to define a generative model $p(\xs \mid \ys)$ which can synthesize new samples $\xs \in \R^D$ while maintaining a link to the classification task solved by TabPFN.
%Here, $K$ represents the number of classes, and $D$ signifies the dimensionality of the data.

TabPFN, like many classification models, induces a conditional distribution $p(y \mid x)$ given by 
$\sigma(f(x))[y]$,
% $\frac{\exp(f(x)[y])}{\sum_{y'}\exp(f(x)[y'])}$, 
where $x$ is the network input, $y$ is the label for $x$, $f: \R^D \rightarrow \R^K$ represents the TabPFN, $\sigma$ is the softmax function, and $[y]$ denotes an indexing operation. The training data $(\xtr, \ytr)$ is also passed to TabPFN, but we omit it in the notation for simplicity.

Next, recalling Bayes' rule, we have $p(x \mid y) \propto p(y \mid x) \cdot p(x)$.
% We thus only need to specify $p(x)$ to fully specify $p(x \mid y)$. %a marginal distribution in $x$ to fully specify the desired conditional. In order to achieve that,
To specify $p(x)$ we take inspiration from \citet{grathwohl2020your} and employ an energy function $E(x)$ -- termed the \emph{class-agnostic energy} -- so that $p(x) \propto \exp(-E(x))$:
\begin{equation}
    E(x) = -\logsumexp_{y'}(f(x)[y'])
\end{equation}
The class-agnostic energy also closely resembles the energy score introduced in \citet{liu2020energy}.
% However, the absence of class-specific information %in class-agnostic energy
%  is insufficient for class-conditional generation.
However, since we are interested in class-\emph{conditional} generation, we also require \emph{class-specific} information.
% Therefore, we further utilize
To do this, first rewrite $p(y \mid x)$ as 
\begin{equation}
    p(y \mid x) = \frac{\exp\left(f(x)[y]\right)}{\sum_{y'} \exp\left(f(x)[y']\right)} = \exp\left(f(x)[y] - \logsumexp_{y'}(f(x)[y'])\right),
\end{equation}
and thus $p(x \mid y) \propto p(y \mid x) \cdot p(x) \propto \exp(f(x)[y])$, as the $\logsumexp$ terms cancel out.
As a result, the \emph{class-conditional energy} can be defined as simply:
\begin{equation}\label{eq:cc-energy}
    E(x \mid y) \coloneqq -f(x)[y]
\end{equation}
We have thus arrived at an energy-based class-conditional generative model $p(x \mid y)$ which relies on TabPFN in a principled manner. % We refer to this generative model as \emph{TabPFGen\textsubscript{core}}.
As depicted in Figure \ref{fig:tabpfgen}, we first obtain $f(\xs)$ using $(\xtr, \ytr)$ as training data. Then, we can directly compute %the class-conditional energy
\eqref{eq:cc-energy} using $E(\xs \mid \ys) = -f(\xs)[\ys]$, eliminating the need to calculate the class-agnostic energy and cross entropy separately.

\begin{algorithm}[h]
\caption{TabPFGen\textsubscript{core}: Given TabPFN model $f$, SGLD step size $\alpha$, SGLD noise $\sigma$, SGLD steps $\eta$, manually defined synthetic labels $\ys$}
\small
\label{alg:tabpfgen_algorithm}
\begin{algorithmic}[1]
\setstretch{1.3} % Adjust the factor to control line spacing
\STATE \textbf{Input:} $\xtr$, $\ytr$
\STATE Initialize $\xs^0$  \hfill$\triangleright$ SGLD Initialization; detailed in Appendix \ref{section:tabpfgen_details}
\FOR{$t \in $[1, 2, ..., $\eta$]}
    \STATE $E(\xs^t \mid \ys) = - f(\xs^t \mid (\xtr, \ytr))[\ys]$ \hfill$\triangleright$ Class-conditional Energy
    \STATE $\xs^{t+1} = \xs^t -\alpha \cdot \frac{\partial E(\xs^t|\ys)}{\partial \xs^t} + \sigma \cdot \mathcal{N}(0, I)$ \hfill$\triangleright$ SGLD
\ENDFOR
\STATE \textbf{Output:} $\xs^\eta$
\end{algorithmic}
\end{algorithm}

% \sim \mathcal{N}(\mathbf{\mu}_{\xtr}, \mathbf{\Sigma}_{\xtr})$

In order to sample from this energy-based model, we employ the stochastic gradient Langevin dynamics (SGLD) method \citep{welling2011bayesian}.
We are able to generate a batch of $\xs$ every $\eta$ steps, taking advantage of their independence.
The choice of $\eta$ is determined based on the classification performance on $\xtr$ using a frozen TabPFN.
This method is referred to as TabPFGen\textsubscript{core} and is detailed in Algorithm \ref{alg:tabpfgen_algorithm}. 

To further enhance the core methodology, we propose an alternative configuration by switching $(\xs, \ys)$ and $(\xtr, \ytr)$ in e.g.\ Figure \ref{fig:tabpfgen}, and integrating the resulting class-conditional energy $E(\xtr \mid \ytr)$ -- which implicitly depends on $(\xs, \ys)$ -- into $E(\xs \mid \ys)$. We posit this modification introduces a regularization effect, although we leave the proof of this to future work. Empirically, this extended approach exhibits slightly superior performance and greater stability. An ablation analysis can be found in Appendix \ref{section: ablation}. The configuration details are detailed in Appendix \ref{section:tabpfgen_details}. We henceforth denote the full approach as \emph{TabPFGen}.

\section{Experiments \& Analysis}
\label{experiments}

We conduct a comprehensive set of experiments utilizing the 18 numerical datasets without missing values, which are previouly employed by \citet{hollmann2023tabpfn}. These datasets are obtained from the OpenML-CC18 suite \citep{bischl2021openml} and their specifics are outlined in Appendix \ref{section:dataset}.
% Our investigation focuses primarily on using synthetic data as augmentation and class balancing. Additionally, we explore the usage for imputation.

\textbf{Experimental Setup:} We partition each dataset into equal-sized training and test sets, and then train or utilize generative models to synthesize samples given the training data. To evaluate the efficacy of a generative model we use its synthetic samples to either replace, augment, or class-balance the training data, then train a variety of discriminative models to predict the class label and evaluate the AUC (Area Under the ROC Curve) performance on the test set. We present the mean and standard deviation of the above evaluation process over three runs with unique seeds.
% Subsequently, we obtain the augmented data by combining the synthetic data with the training data. Finally, 

\textbf{Downstream Models:} For all experiments we train and evaluate 4 distinct downstream models: \textit{XGBoost} \citep{chen2016xgboost}, \textit{random forest} \citep{ho1995}, \textit{logistic regression}, and \textit{TabPFN} \citep{hollmann2023tabpfn}.
To ensure fair comparisons we adopt the optimized hyperparameters of downstream models previously published by \citet{hollmann2023tabpfn}, if available. Experiments with a range of alternative hyperparameter configurations consistently demonstrate that TabPFGen outperforms baseline methods, as detailed in Appendix \ref{section:downstream_model_hyperparameters}.
% Moreover, we conduct experiments with a range of alternative hyperparameters, as detailed in Appendix \ref{section:downstream_model_hyperparameters}. Our findings consistently demonstrate that TabPFGen outperforms baseline methods across various downstream model hyperparameter configurations.

% Moreover, we conduct experiments with a range of alternative hyperparameters, as detailed in Appendix \ref{section:downstream_model_hyperparameters}. Our findings consistently demonstrate that TabPFGen outperforms baseline methods across various downstream model hyperparameter configurations.

\textbf{Baseline Models:} We employ a diverse set of baseline generative models, including the traditional interpolation approach SMOTE \citep{chawla2002smote}, generative adversarial networks represented by CTGAN \citep{xu2019modeling}, variational autoencoder-based methods including TVAE \citep{xu2019modeling} and RTVAE \citep{akrami2020robust}, the normalizing flow-based Neural Spline Flows (NF) \citep{durkan2019neural}, and the recent diffusion-based methods TabDDPM \citep{kotelnikov2023tabddpm}.
All baseline generative models are trained and evaluated using the publicly available \texttt{synthcity} package \citep{qian2023synthcity}.
The details and hyperparameters of each generative model can be found in Appendix \ref{section:baselines}.

\subsection{Results}

\textbf{Replacement \& Augmentation:} 
We assess the quality of synthetic samples from a given model through two tasks: \textit{augmentation}, which augments the training dataset with an equal volume and class-ratio of synthetic data, and \textit{replacement}, which replaces the original training set with synthetic data.
The replacement task assesses whether the generated samples closely resemble the training data, while the augmentation task assesses whether adding synthetic samples improves classification performance.
Table \ref{tab:augmentation_table} displays the downstream model performance on the test sets when utilizing synthetic samples from various generative models during training. We find that TabPFGen facilitates a significant performance enhancement over other generative models -- TabPFGen is the only generative model that consistently improves downstream performance when augmenting the original training set. For instance, with XGBoost, the average AUC (0.936) achieved is better than a model trained on the original data (0.924). Individual results on each of the 18 datasets can be found in Appendix \ref{section:details_synthetic_data_augmentation}. TabPFGen also outperforms all other generative methods on the replacement task, indicating that its generated samples most closely resemble the training data. We highlight that TabPFGen requires no training or fine-tuning, and no hyperparameter tuning to adapt to specific datasets.

% \vspace{-1em}
% \vspace{-6pt}
% \vspace{-10pt}
\begin{table}[h]
  \caption{Average AUC over OpenML-CC18 test sets, with error bars over 3 runs. Top 4 rows: synthetic data as augmentation. Bottom 4 rows: synthetic data as replacement.}
  \label{tab:augmentation_table}
  \centering
  \vspace{-5pt}
  \scriptsize % Smaller font size for the table
  \setlength{\tabcolsep}{3.6pt} % Reduce the column padding by half
  \begin{tabular}{@{}lcccccccc@{}}
    \toprule
    Model & Original & \ttfamily{SMOTE} & \ttfamily{CTGAN} & \ttfamily{TVAE} & \ttfamily{NF} & \ttfamily{RTVAE} & \ttfamily{TabDDPM} & \ttfamily{TabPFGen} \\
    \midrule
    XGB 
    & $0.924 \scriptscriptstyle \pm 3e-4$ 
    & $0.926 \scriptscriptstyle \pm 3e-4$ 
    & $0.912 \scriptscriptstyle \pm 2e-4$ 
    & $0.914 \scriptscriptstyle \pm 7e-4$ 
    & $0.912 \scriptscriptstyle \pm 4e-4$ 
    & $0.917 \scriptscriptstyle \pm 3e-4$ 
    & $0.927 \scriptscriptstyle \pm 3e-4$ 
    & $\mathbf{0.934} \scriptscriptstyle \pm 3e-4$ \\
    
    RF 
    & $0.906 \scriptscriptstyle \pm 3e-4$ 
    & $0.906 \scriptscriptstyle \pm 2e-3$  
    & $0.898 \scriptscriptstyle \pm 1e-3$ 
    & $0.904 \scriptscriptstyle \pm 1e-3$ 
    & $0.894 \scriptscriptstyle \pm 3e-4$ 
    & $0.907 \scriptscriptstyle \pm 2e-3$  
    & $0.911 \scriptscriptstyle \pm 7e-4$ 
    & $\mathbf{0.912} \scriptscriptstyle \pm 4e-4$ \\
    
    LR 
    & $0.920 \scriptscriptstyle \pm 7e-4$ 
    & $0.914 \scriptscriptstyle \pm 3e-3$ 
    & $0.904 \scriptscriptstyle \pm 3e-3$ 
    & $0.909 \scriptscriptstyle \pm 6e-3$ 
    & $0.901 \scriptscriptstyle \pm 9e-4$ 
    & $0.906 \scriptscriptstyle \pm 8e-3$  
    & $0.885 \scriptscriptstyle \pm 3e-4$ 
    & $\mathbf{0.921} \scriptscriptstyle \pm 2e-4$ \\
    
    TabPFN 
    & $0.934 \scriptscriptstyle \pm 2e-3$ 
    & $0.927 \scriptscriptstyle \pm 1e-3$ 
    & $0.930 \scriptscriptstyle \pm 1e-3$ 
    & $0.931 \scriptscriptstyle \pm 1e-3$ 
    & $0.928 \scriptscriptstyle \pm 3e-4$ 
    & $0.932 \scriptscriptstyle \pm 1e-3$ 
    & $0.929 \scriptscriptstyle \pm 5e-4$ 
    & $\mathbf{0.935} \scriptscriptstyle \pm 3e-4$ \\

    \midrule
    
    XGB 
    & N/A
    & $0.907 \scriptscriptstyle \pm 4e-4$ 
    & $0.842 \scriptscriptstyle \pm 8e-4$
    & $0.858 \scriptscriptstyle \pm 2e-3$
    & $0.700 \scriptscriptstyle \pm 6e-4$
    & $0.795 \scriptscriptstyle \pm 9e-4$
    & $0.812 \scriptscriptstyle \pm 3e-4$
    & $\mathbf{0.927} \scriptscriptstyle \pm 3e-4$ \\
    
    RF 
    & N/A
    & $0.894 \scriptscriptstyle \pm 1e-3$ 
    & $0.837 \scriptscriptstyle \pm 6e-4$
    & $0.844 \scriptscriptstyle \pm 5e-4$
    & $0.676 \scriptscriptstyle \pm 2e-3$
    & $0.774 \scriptscriptstyle \pm 3e-4$
    & $0.814 \scriptscriptstyle \pm 9e-4$
    & $\mathbf{0.906} \scriptscriptstyle \pm 6e-4$ \\
    
    LR 
    & N/A
    & $0.893 \scriptscriptstyle \pm 2e-3$ 
    & $0.843 \scriptscriptstyle \pm 6e-4$
    & $0.873 \scriptscriptstyle \pm 1e-3$
    & $0.722 \scriptscriptstyle \pm 3e-3$
    & $0.854 \scriptscriptstyle \pm 7e-4$
    & $0.876 \scriptscriptstyle \pm 3e-4$
    & $\mathbf{0.920} \scriptscriptstyle \pm 1e-3$ \\
    
    TabPFN 
    & N/A
    & $0.920 \scriptscriptstyle \pm 8e-4$ 
    & $0.888 \scriptscriptstyle \pm 4e-4$
    & $0.887 \scriptscriptstyle \pm 3e-4$
    & $0.705 \scriptscriptstyle \pm 2e-3$
    & $0.862 \scriptscriptstyle \pm 1e-3$
    & $0.894 \scriptscriptstyle \pm 7e-4$
    & $\mathbf{0.934} \scriptscriptstyle \pm 2e-4$ \\
    \bottomrule
  \end{tabular}
\end{table}
\begin{table}[ht!]
  \caption{Average AUC using synthetic data for class balancing. Error bars can be found in the Appendix. These datasets are the five most class-imbalanced datasets in the OpenML-CC18 suite.}
  \label{tab:class_balancing}
  \centering
  \vspace{-5pt}
  \scriptsize % Smaller font size for the table
  \setlength{\tabcolsep}{5pt} % Reduce the column padding by half
  \begin{tabular}{@{}lcccccccccc@{}}
    \toprule
    Dataset & Original & Sampling & SMOTE & CTGAN & TVAE & NF & RTVAE & TabDDPM & TabPFGen \\
    \midrule
    KC & $0.823$ & $0.875$ & $0.872$ & $0.859$ & $0.848$ & $0.862$ & $0.866$ & $0.805$ & $\mathbf{0.877}$ \\ % i2 adam blend 0.9 frac
    PC & $0.824$ & $0.811$ & $0.836$ & $0.827$ & $0.835$ & $0.825$ & $\mathbf{0.841}$ & $0.825$ & $\mathbf{0.841}$ \\ % i4 sgd tg 0.9 frac
    BL & $0.731$ & $0.757$ & $0.756$ & $0.743$ & $0.714$ & $0.755$ & $0.706$ & $\mathbf{0.774}$ & $0.767$ \\
    CL & $0.925$ & $0.935$ & $0.949$ & $0.793$ & $0.771$ & $0.795$ & $0.909$ & $0.915$ & $\mathbf{0.955}$ \\ % i2 adam blend 0.9 frac
    DI & $0.837$ & $0.832$ & $0.832$ & $0.831$ & $0.813$ & $0.832$ & $0.837$ & $0.843$ & $\mathbf{0.844}$ \\ % i2 adam blend 0.9 frac
    \bottomrule
  \end{tabular}
\end{table}

% \textbf{Synthetic Data as Augmentation:} To use synthetic data for augmentation, we combine the training dataset with an equal volume of synthetic data, preserving the original class ratio. The top 4 rows in Table \ref{tab:augmentation_table} present the original downstream model's performance (second column) and the performance on the augmented data using various techniques (subsequent columns). Our findings reveal a significant performance enhancement facilitated by TabPFGen. For instance, in the case of XGBoost, the new AUC (0.936) obtained is better than the previous state-of-the-art (0.934), and notably, TabPFN performance further improves to 0.937 with the inclusion of TabPFGen as well. Detailed results can be found in Table \ref{tab:experiment_details_augment}. Furthermore, we analyze the performance of downstream models using \textit{only} synthetic data. As shown in the bottom 4 rows in Table \ref{tab:augmentation_table}, TabPFGen still consistently outperforms other competitive baselines.

% \vspace{-16pt}
% analysis outline:
% \begin{itemize}
%     \item two moon density plot for CTGAN, TabDDPM, TabPFGen
%     \item ablation studies for Train-Synth, Synth-Train, Blend and top-k
% \end{itemize}
% \subsection{Synthetic Data Quality Analysis}
% \vspace{-10pt}
\begin{figure}[ht!]
    \centering
    \begin{subfigure}{0.19\textwidth}
        \centering
        \includegraphics[width=\linewidth]{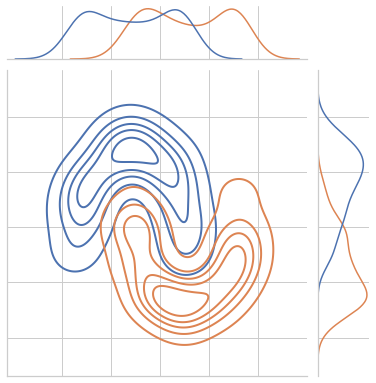}
        \caption{Original}
    \end{subfigure}
    \begin{subfigure}{0.19\textwidth}
        \centering
        \includegraphics[width=\linewidth]{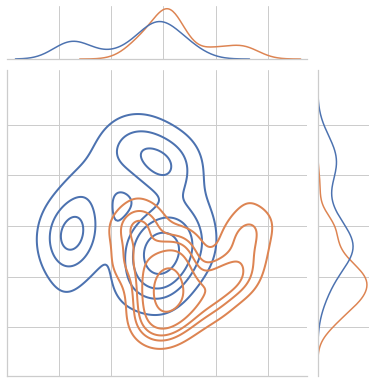}
        \caption{CTGAN}
    \end{subfigure}
    \begin{subfigure}{0.19\textwidth}
        \centering
        \includegraphics[width=\linewidth]{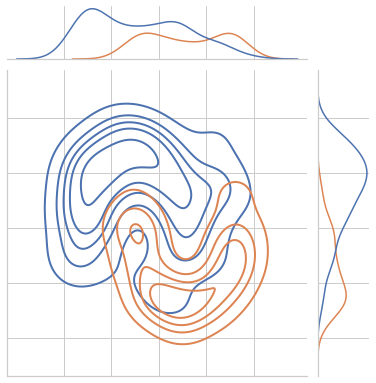}
        \caption{NF}
    \end{subfigure}
    \begin{subfigure}{0.19\textwidth}
        \centering
        \includegraphics[width=\linewidth]{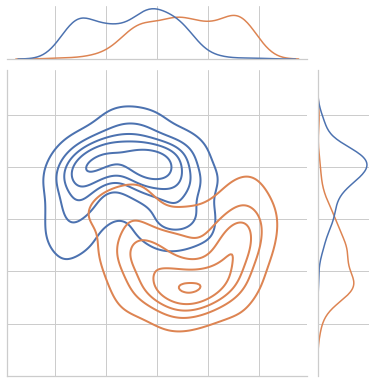}
        \caption{TabDDPM}
    \end{subfigure}
    \begin{subfigure}{0.19\textwidth}
        \centering
        \includegraphics[width=\linewidth]{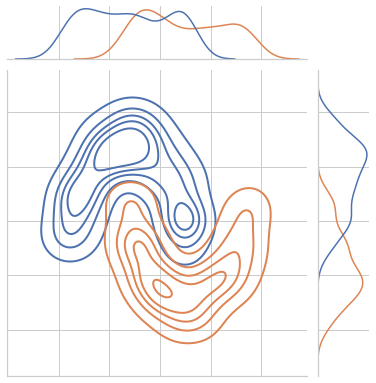}
        \caption{TabPFGen}
    \end{subfigure}
    \caption{Contour and marginal density plots of: (a) original two-moons dataset; (b)-(d) synthetic data generated using baseline methods; (e) synthetic data generated by TabPFGen}
    \label{fig:qualitative_analysis}
    % \vspace{-5mm}
\end{figure}

\textbf{Class Balancing:} We create class-balanced datasets by generating synthetic data for minority classes until every class has an equal number of samples.
We then train an XGBoost model on the class-balanced data, and report its test AUC in Table \ref{tab:class_balancing}.
We find that TabPFGen generally outperforms alternative methods, including even class-balancing the original data by sampling with replacement, which we denote ``sampling''. Additional details and an expanded table can be found in Appendix~\ref{section:additional_class_balancing}.

% To perform class balancing, we keep the amount of data in the majority class fixed and only generate synthetic data for the minority classes such that every class has equal amount of data. We then train an XGBoost model using the class balanced data and report its validation AUC. As shown in Table \ref{tab:class_balancing}, we find TabPFGen generally performs better than other known methods. Here, sampling represents re-sampling the original data with replacement. Additional details of the class balancing experiment and the dataset description can be found in Tables \ref{tab:class_balancing_std} and \ref{tab:class_balancing_datasets} in the appendix.

\textbf{Imputation:} Promising results on imputation using TabPFGen are shown in Appendix \ref{section:imputation}.

\textbf{Qualitative Assessment:} The above analyses demonstrate that TabPFN's generated samples have a high degree of utility for downstream applications. As a further confirmation of sample quality we display contour and the marginal plots for generative models on the popular two-moons dataset in Figure \ref{fig:qualitative_analysis}. We show that the synthetic data distribution generated by TabPFGen is more similar to the original data distribution than other baseline methods. Surprisingly, we find that some generative models perform poorly even on such a simple dataset.

% To further analyze the quality of the synthetic data, we use the popular two-moons dataset as the training set and generate synthetic data using TabPFGen and other baseline methods. As shown in Figure \ref{fig:qualitative_analysis}, both the contour and the marginal plots indicate that the synthetic data distribution generated by TabPFGen is more similar to the original data distribution than other baseline methods.
\section{Conclusion}
In this work, we present TabPFGen, an efficient approach harnessing a pre-trained TabPFN as an energy based model for tabular data generation without additional training. TabPFGen outperforms competitive baselines across augmentation, class balancing, and imputation tasks. Nonetheless, it is essential to recognize that TabPFN's current limitations, such as input size constraints and a focus on numerical datasets, limits our method's current applicability for large-scale datasets. We continue this discussion in Appendix \ref{section:limitations_and_impact}, but anticipate that these limitations will gradually recede as transformers and extensions of TabPFN advance, and argue that re-purposing powerful pretrained discriminative models for tabular generation is an impactful and promising avenue for future research.

\bibliographystyle{plainnat}
\bibliography{refs}

\begin{thebibliography}{42}
\providecommand{\natexlab}[1]{#1}
\providecommand{\url}[1]{\texttt{#1}}
\expandafter\ifx\csname urlstyle\endcsname\relax
  \providecommand{\doi}[1]{doi: #1}\else
  \providecommand{\doi}{doi: \begingroup \urlstyle{rm}\Url}\fi

\bibitem[san(2019)]{santurkar2019image}
Image synthesis with a single (robust) classifier.
\newblock In \emph{Santurkar, Shibani and Ilyas, Andrew and Tsipras, Dimitris
  and Engstrom, Logan and Tran, Brandon and Madry, Aleksander}, 2019.

\bibitem[Akrami et~al.(2022)Akrami, Joshi, Li, Aydöre, and
  Leahy]{akrami2020robust}
Haleh Akrami, Anand~A. Joshi, Jian Li, Sergül Aydöre, and Richard~M. Leahy.
\newblock A robust variational autoencoder using beta divergence.
\newblock \emph{Knowledge-Based Systems}, 238:\penalty0 107886, 2022.

\bibitem[Benjelloun et~al.(2020)Benjelloun, Chen, and
  Noy]{benjelloun2020google}
Omar Benjelloun, Shiyu Chen, and Natasha Noy.
\newblock Google dataset search by the numbers.
\newblock In \emph{The Semantic Web -- ISWC 2020}, pages 667--682, 2020.

\bibitem[Bischl et~al.(2021)Bischl, Casalicchio, Feurer, Gijsbers, Hutter,
  Lang, Gomes~Mantovani, van Rijn, and Vanschoren]{bischl2021openml}
Bernd Bischl, Giuseppe Casalicchio, Matthias Feurer, Pieter Gijsbers, Frank
  Hutter, Michel Lang, Rafael Gomes~Mantovani, Jan van Rijn, and Joaquin
  Vanschoren.
\newblock {OpenML} benchmarking suites.
\newblock In \emph{Proceedings of the Neural Information Processing Systems
  Track on Datasets and Benchmarks}, 2021.

\bibitem[Borisov et~al.(2022)Borisov, Sessler, Leemann, Pawelczyk, and
  Kasneci]{borisov2022language}
Vadim Borisov, Kathrin Sessler, Tobias Leemann, Martin Pawelczyk, and Gjergji
  Kasneci.
\newblock Language models are realistic tabular data generators.
\newblock In \emph{International Conference on Learning Representations}, 2022.

\bibitem[Brock et~al.(2018)Brock, Donahue, and Simonyan]{brock2018large}
Andrew Brock, Jeff Donahue, and Karen Simonyan.
\newblock Large scale {GAN} training for high fidelity natural image synthesis.
\newblock In \emph{International Conference on Learning Representations}, 2018.

\bibitem[Camino et~al.(2020)Camino, State, and
  Hammerschmidt]{pmlr-v137-camino20a}
Ramiro~D. Camino, Radu State, and Christian~A. Hammerschmidt.
\newblock Oversampling tabular data with deep generative models: Is it worth
  the effort?
\newblock In \emph{Proceedings on "I Can't Believe It's Not Better!" at NeurIPS
  Workshops}, pages 148--157, 2020.

\bibitem[Chawla et~al.(2002)Chawla, Bowyer, Hall, and
  Kegelmeyer]{chawla2002smote}
Nitesh~V Chawla, Kevin~W Bowyer, Lawrence~O Hall, and W~Philip Kegelmeyer.
\newblock {SMOTE}: Synthetic minority over-sampling technique.
\newblock \emph{Journal of Artificial Intelligence Research}, 16:\penalty0
  321--357, 2002.

\bibitem[Chen and Guestrin(2016)]{chen2016xgboost}
Tianqi Chen and Carlos Guestrin.
\newblock {XGBoost}: A scalable tree boosting system.
\newblock In \emph{Proceedings of the 22nd ACM SigKDD International Conference
  on Knowledge Discovery and Data Mining}, pages 785--794, 2016.

\bibitem[Clements et~al.(2020)Clements, Xu, Yousefi, and
  Efimov]{clements2020sequential}
Jillian~M Clements, Di~Xu, Nooshin Yousefi, and Dmitry Efimov.
\newblock Sequential deep learning for credit risk monitoring with tabular
  financial data.
\newblock \emph{arXiv preprint arXiv:2012.15330}, 2020.

\bibitem[Devlin et~al.(2019)Devlin, Chang, Lee, and Toutanova]{devlin2018bert}
Jacob Devlin, Ming-Wei Chang, Kenton Lee, and Kristina Toutanova.
\newblock {BERT}: Pre-training of deep bidirectional transformers for language
  understanding.
\newblock In \emph{Proceedings of the 2019 Conference of the North American
  Chapter of the Association for Computational Linguistics: Human Language
  Technologies}, pages 4171--4186, 2019.

\bibitem[Du and Mordatch(2019)]{du2019implicit}
Yilun Du and Igor Mordatch.
\newblock Implicit generation and modeling with energy based models.
\newblock In \emph{Advances in Neural Information Processing Systems}, 2019.

\bibitem[Durkan et~al.(2019)Durkan, Bekasov, Murray, and
  Papamakarios]{durkan2019neural}
Conor Durkan, Artur Bekasov, Iain Murray, and George Papamakarios.
\newblock Neural spline flows.
\newblock \emph{Advances in Neural Information Processing systems}, 2019.

\bibitem[Engelmann and Lessmann(2021)]{engelmann2021conditional}
Justin Engelmann and Stefan Lessmann.
\newblock Conditional {Wasserstein} {GAN-based} oversampling of tabular data
  for imbalanced learning.
\newblock \emph{Expert Systems with Applications}, 174:\penalty0 114582, 2021.

\bibitem[Florence et~al.(2022)Florence, Lynch, Zeng, Ramirez, Wahid, Downs,
  Wong, Lee, Mordatch, and Tompson]{florence2022implicit}
Pete Florence, Corey Lynch, Andy Zeng, Oscar~A Ramirez, Ayzaan Wahid, Laura
  Downs, Adrian Wong, Johnny Lee, Igor Mordatch, and Jonathan Tompson.
\newblock Implicit behavioral cloning.
\newblock In \emph{Conference on Robot Learning}, pages 158--168, 2022.

\bibitem[Gao et~al.(2017)Gao, Lu, Zhou, Zhu, and Wu]{gao2017learning}
Ruiqi Gao, Yang Lu, Junpei Zhou, Song-Chun Zhu, and Ying~Nian Wu.
\newblock Learning energy-based models as generative {ConvNets} via multi-grid
  modeling and sampling.
\newblock \emph{arXiv preprint arXiv:1709.08868}, 2017.

\bibitem[Grathwohl et~al.(2020)Grathwohl, Wang, Jacobsen, Duvenaud, Norouzi,
  and Swersky]{grathwohl2020your}
Will Grathwohl, Kuan-Chieh Wang, Joern-Henrik Jacobsen, David Duvenaud,
  Mohammad Norouzi, and Kevin Swersky.
\newblock Your classifier is secretly an energy based model and you should
  treat it like one.
\newblock In \emph{International Conference on Learning Representations}, 2020.

\bibitem[Ho(1995)]{ho1995}
Tin~Kam Ho.
\newblock Random decision forests.
\newblock In \emph{Proceedings of 3rd International Conference on Document
  Analysis and Recognition}, pages 278--282, 1995.

\bibitem[Hollmann et~al.(2023)Hollmann, M{\"u}ller, Eggensperger, and
  Hutter]{hollmann2023tabpfn}
Noah Hollmann, Samuel M{\"u}ller, Katharina Eggensperger, and Frank Hutter.
\newblock {TabPFN}: A transformer that solves small tabular classification
  problems in a second.
\newblock In \emph{International Conference on Learning Representations}, 2023.

\bibitem[Jordon et~al.(2018)Jordon, Yoon, and Van Der~Schaar]{jordon2018pate}
James Jordon, Jinsung Yoon, and Mihaela Van Der~Schaar.
\newblock {PATE-GAN}: Generating synthetic data with differential privacy
  guarantees.
\newblock In \emph{International Conference on Learning Representations}, 2018.

\bibitem[Kotelnikov et~al.(2023)Kotelnikov, Baranchuk, Rubachev, and
  Babenko]{kotelnikov2023tabddpm}
Akim Kotelnikov, Dmitry Baranchuk, Ivan Rubachev, and Artem Babenko.
\newblock {TabDDPM}: Modelling tabular data with diffusion models.
\newblock In \emph{International Conference on Machine Learning}, pages
  17564--17579, 2023.

\bibitem[Li(2022)]{li2022use}
Haoyang Li.
\newblock Use classifier as generator.
\newblock \emph{arXiv preprint arXiv:2209.09210}, 2022.

\bibitem[Liu et~al.(2020)Liu, Wang, Owens, and Li]{liu2020energy}
Weitang Liu, Xiaoyun Wang, John Owens, and Yixuan Li.
\newblock Energy-based out-of-distribution detection.
\newblock In \emph{Advances in Neural Information Processing Systems}, pages
  21464--21475, 2020.

\bibitem[Ma et~al.(2019)Ma, Zhou, Li, Neubig, and Hovy]{ma2019flowseq}
Xuezhe Ma, Chunting Zhou, Xian Li, Graham Neubig, and Eduard Hovy.
\newblock {FlowSeq}: Non-autoregressive conditional sequence generation with
  generative flow.
\newblock In \emph{Proceedings of the 2019 Conference on Empirical Methods in
  Natural Language Processing and the 9th International Joint Conference on
  Natural Language Processing}, pages 4282--4292, 2019.

\bibitem[Manousakas and Ayd{\"o}re(2023)]{manousakas2023usefulness}
Dionysis Manousakas and Serg{\"u}l Ayd{\"o}re.
\newblock On the usefulness of synthetic tabular data generation.
\newblock In \emph{Data-centric Machine Learning Research (DMLR) Workshop at
  the 40th International Conference on Machine Learning}, 2023.

\bibitem[More(2016)]{more2016survey}
Ajinkya More.
\newblock Survey of resampling techniques for improving classification
  performance in unbalanced datasets.
\newblock \emph{arXiv preprint arXiv:1608.06048}, 2016.

\bibitem[M{\"u}ller et~al.(2022)M{\"u}ller, Hollmann, Arango, Grabocka, and
  Hutter]{muller2022transformers}
Samuel M{\"u}ller, Noah Hollmann, Sebastian~Pineda Arango, Josif Grabocka, and
  Frank Hutter.
\newblock Transformers can do {Bayesian} inference.
\newblock In \emph{International Conference on Learning Representations}, 2022.

\bibitem[Nock and Guillame-Bert(2022)]{nock2022generative}
Richard Nock and Mathieu Guillame-Bert.
\newblock Generative trees: Adversarial and copycat.
\newblock In \emph{International Conference on Machine Learning}, pages
  16906--16951, 2022.

\bibitem[Qian et~al.(2023)Qian, Cebere, and van~der Schaar]{qian2023synthcity}
Zhaozhi Qian, Bogdan-Constantin Cebere, and Mihaela van~der Schaar.
\newblock Synthcity: Facilitating innovative use cases of synthetic data in
  different data modalities.
\newblock \emph{arXiv preprint arXiv:2301.07573}, 2023.

\bibitem[Rombach et~al.(2022)Rombach, Blattmann, Lorenz, Esser, and
  Ommer]{rombach2022high}
Robin Rombach, Andreas Blattmann, Dominik Lorenz, Patrick Esser, and Bj{\"o}rn
  Ommer.
\newblock High-resolution image synthesis with latent diffusion models.
\newblock In \emph{Proceedings of the IEEE/CVF Conference on Computer Vision
  and Pattern Recognition}, pages 10684--10695, 2022.

\bibitem[Shwartz-Ziv and Armon(2022)]{shwartz2022tabular}
Ravid Shwartz-Ziv and Amitai Armon.
\newblock Tabular data: Deep learning is not all you need.
\newblock \emph{Information Fusion}, 81:\penalty0 84--90, 2022.

\bibitem[Solatorio and Dupriez(2023)]{solatorio2023realtabformer}
Aivin~V. Solatorio and Olivier Dupriez.
\newblock {REaLTabFormer}: Generating realistic relational and tabular data
  using transformers.
\newblock \emph{arXiv preprint arXiv:2302.02041}, 2023.

\bibitem[Tang et~al.(2020)Tang, Xia, Zhang, and Long]{tang2020customer}
Qi~Tang, Guoen Xia, Xianquan Zhang, and Feng Long.
\newblock A customer churn prediction model based on {XGBoost and MLP}.
\newblock In \emph{2020 International Conference on Computer Engineering and
  Application (ICCEA)}, pages 608--612, 2020.

\bibitem[Tu(2007)]{tu2007learning}
Zhuowen Tu.
\newblock Learning generative models via discriminative approaches.
\newblock In \emph{Proceedings of the IEEE/CVF Conference on Computer Vision
  and Pattern Recognition}, pages 1--8, 2007.

\bibitem[Ulmer et~al.(2020)Ulmer, Meijerink, and Cin{\`a}]{ulmer2020trust}
Dennis Ulmer, Lotta Meijerink, and Giovanni Cin{\`a}.
\newblock Trust issues: Uncertainty estimation does not enable reliable {OOD}
  detection on medical tabular data.
\newblock In \emph{Machine Learning for Health}, pages 341--354, 2020.

\bibitem[Urban and Gates(2021)]{urban2021deep}
Christopher~J Urban and Kathleen~M Gates.
\newblock Deep learning: A primer for psychologists.
\newblock \emph{Psychological Methods}, 26\penalty0 (6):\penalty0 743, 2021.

\bibitem[Vahdat and Kautz(2020)]{vahdat2020nvae}
Arash Vahdat and Jan Kautz.
\newblock {NVAE}: A deep hierarchical variational autoencoder.
\newblock In \emph{Advances in Neural Information Processing Systems}, 2020.

\bibitem[Welling and Teh(2011)]{welling2011bayesian}
Max Welling and Yee~Whye Teh.
\newblock Bayesian learning via stochastic gradient {L}angevin dynamics.
\newblock In \emph{International Conference on Machine Learning}, pages
  681--688, 2011.

\bibitem[Xie et~al.(2016)Xie, Lu, Zhu, and Wu]{xie2016theory}
Jianwen Xie, Yang Lu, Song-Chun Zhu, and Yingnian Wu.
\newblock A theory of generative {ConvNet}.
\newblock In \emph{International Conference on Machine Learning}, pages
  2635--2644, 2016.

\bibitem[Xu et~al.(2019)Xu, Skoularidou, Cuesta-Infante, and
  Veeramachaneni]{xu2019modeling}
Lei Xu, Maria Skoularidou, Alfredo Cuesta-Infante, and Kalyan Veeramachaneni.
\newblock Modeling tabular data using conditional {GAN}.
\newblock In \emph{Advances in Neural Information Processing Systems}, 2019.

\bibitem[Yin et~al.(2021)Yin, Mallya, Vahdat, Alvarez, Kautz, and
  Molchanov]{yin2021see}
Hongxu Yin, Arun Mallya, Arash Vahdat, Jose~M Alvarez, Jan Kautz, and Pavlo
  Molchanov.
\newblock See through gradients: Image batch recovery via {GradInversion}.
\newblock In \emph{Proceedings of the IEEE/CVF Conference on Computer Vision
  and Pattern Recognition}, pages 16337--16346, 2021.

\bibitem[Zheng and Charoenphakdee(2022)]{zheng2022diffusion}
Shuhan Zheng and Nontawat Charoenphakdee.
\newblock Diffusion models for missing value imputation in tabular data.
\newblock In \emph{NeurIPS 2022 First Table Representation Workshop}, 2022.

\end{thebibliography}

%%%%%%%%%%%%%%%%%%%%%%%%%%%%%%%%%%%%%%%%%%%%%%%%%%%%%%%%%%%%

\clearpage

\appendix

\section{Limitations and Impact}
\label{section:limitations_and_impact}

Our method builds upon TabPFN, and consequently inherits certain limitations.
Presently, TabPFN imposes restrictions on input size, allowing for a maximum of 2000 tokens, 100 features, and 10 classes. These constraints arise from the quadratic complexity inherent in the underlying transformer architecture. It is important to note that TabPFN and TabPFGen algorithms are designed to be architecture-agnostic, therefore the improvement of transformer architectures will directly translate to our method. As the research of transformer architectures continues to advance, we anticipate these constraints will diminish.

% Additionally, our TabPFGen sampling process involves a number of steps prior to the actual sampling process. This behavior aligns with other SGLD-based methods but may not be ideal. We acknowledge that this aspect can be further refined in future research.

% Finally, although we have a strong theoretical understanding on the workings of the base TabPFGen with SGLD, we have less theoretical motivation for the use of ``Adam plus noise'', as well as the swapping of $(\xs, \ys)$ with $(\xtr, \ytr)$ to augment the loss function within the sampling procedure.
% For the former, we anticipate that leveraging the deep connection between sampling and optimization \citep{cheng2020interplay} could provide some insight into what type of distribution is being sampled from when using ``Adam plus noise''.
% For the latter, empirically we have observed a ``mode-seeking'' effect from building the core TabPFGen exclusively as in Figure \ref{fig:tabpfgen}, whereas we have observed a ``mode-covering'' effect when swapping $(\xs, \ys)$ with $(\xtr, \ytr)$; it is thus not surprising that combining the two would result in even better performance.
% It may be possible to say something concrete using the form of the KL divergence optimized by the original TabPFN \citep{hollmann2023tabpfn}, but this is outside the scope of the current work.
% In either case, the empirical motivation for these extensions was quite clear, and thus we elect to include them despite the lack of a thorough explanation into their performance.

\section{Datasets Details}
\label{section:dataset}
Our experiments are conducted using the 18 numerical datasets from OpenML-CC18 \citep{bischl2021openml} (available at \url{http://openml.org}). Similar to \citet{hollmann2023tabpfn}, we use datasets with maximum 2000 samples, 100 numerical features, and 10 classes, without missing values. The details of the datasets are listed in Table \ref{fig:openml-datasets}.

\begin{table}[h]
    \caption{18 datasets from OpenML-CC18}
    \label{fig:openml-datasets}
    \centering
    \small
    \begin{tabular}{@{\hskip 0mm}l@{\hskip 0mm}rrrrr}
        \toprule
         Name & \#Feat. &  \#Inst. &  \#Class. &  Minor. Class Size & OpenML ID \\
        \midrule
         balance-scale & 5 & 625 & 3 & 49 & 11 \\
         mfeat-fourier & 77 & 2000 & 10 & 200 & 14 \\
         mfeat-karhunen & 65 & 2000 & 10 & 200 & 16 \\
         mfeat-morphological & 7 & 2000 & 10 & 200 & 18 \\
         mfeat-zernike & 48 & 2000 & 10 & 200 & 22 \\
         diabetes & 9 & 768 & 2 & 268 & 37 \\
         vehicle & 19 & 846 & 4 & 199 & 54 \\
         analcatdata\_auth... & 71 & 841 & 4 & 55 & 458 \\
         pc4 & 38 & 1458 & 2 & 178 & 1049 \\
         pc3 & 38 & 1563 & 2 & 160 & 1050 \\
         kc2 & 22 & 522 & 2 & 107 & 1063 \\
         pc1 & 22 & 1109 & 2 & 77 & 1068 \\
         banknote-authenti... & 5 & 1372 & 2 & 610 & 1462 \\
        blood-transfusion-... & 5 & 748 & 2 & 178 & 1464 \\
        qsar-biodeg & 42 & 1055 & 2 & 356 & 1494 \\
        wdbc & 31 & 569 & 2 & 212 & 1510 \\
        steel-plates-fault & 28 & 1941 & 7 & 55 & 40982 \\
        climate-model-simu... & 21 & 540 & 2 & 46 & 40994 \\
        \bottomrule
    \end{tabular}
\end{table}

\section{Experimental Details}
\subsection{Baseline Details}
\label{section:baselines}
We use the \texttt{synthcity} package \citep{qian2023synthcity} to run all of our baselines. The \texttt{synthcity} code repository can be found at: \url{https://github.com/vanderschaarlab/synthcity}. We keep the same hyperparameters as \citet{manousakas2023usefulness} when possible. The exact hyperparameters and training details can be found in Table \ref{tab:baseline_hyperparams}.

\subsection{TabPFGen Details}
\label{section:tabpfgen_details}

We use a pre-trained TabPFN for all of our experiments. The pre-trained TabPFN weights are available for download from the following link: \url{https://github.com/automl/TabPFN/raw/main/tabpfn/models_diff/prior_diff_real_checkpoint_n_0_epoch_42.cpkt}. 

We first initialize $\xs$ using the training data $\xtr$ injected with Gaussian noise. We fix the mean of the noise to be 0 and standard deviation to be 0.01 throughout all of our experiments. We then perform SGLD for $\eta$ steps guided by the classification AUC on $\xtr$ with $\xs$ as in-context training data. Since TabPFN accepts independent inputs, we are able to generate a batch of independent samples for each iteration in the SGLD process. Therefore, we return the best batch of $\xs$ at the end of the SGLD process.

\section{Additional Results}
\subsection{Imputation with TabPFGen}
\label{section:imputation}

\begin{figure}[h]
    % \centering
    \begin{subfigure}[b]{0.18\textwidth}
        \centering
        \begin{tikzpicture}[scale=0.4]
        \begin{axis}[
            title={Balance Scale},
            xlabel={Percentage of Data Missing [\%]},
            ylabel={RMSE},
            xmin=20, xmax=70,
            ymin=0.88, ymax=1.04,
            xtick={70, 60, 50, 40, 30, 20},
            ytick={0.88, 0.90, 0.92, 0.94, 0.96, 0.98, 1.00, 1.02, 1.04},
            legend style={at={(0.98,0.5)},anchor=south east},
            ymajorgrids=true,
            grid style=dashed,
            x dir=reverse
        ]
        
        \addplot[
            color=red,
            mark=square,
            ]
            coordinates {
            (20,0.928)(30,0.933)(40,0.932)(50,0.928)(60,0.95)(70, 0.989)
            };
            
        \addplot[
            color=blue,
            mark=*,
            ]
            coordinates {
            (20,1.04)(30,1.028)(40,1.028)(50,1.014)(60,1.017)(70, 1.018)
            };
        
        \addlegendentry{TabPFGen}
        \addlegendentry{Mean}
            
        \end{axis}
        \end{tikzpicture}
        % \caption{Balance Scale}
    \end{subfigure}%
    % \hfill
    \hspace{0.06\textwidth} % Adjust the spacing here
    \begin{subfigure}[b]{0.18\textwidth}
        \centering
        \begin{tikzpicture}[scale=0.4]
        \begin{axis}[
            title={Mfeat Morphological},
            xlabel={Percentage of Data Missing [\%]},
            ylabel={RMSE},
            xmin=20, xmax=70,
            ymin=0.78, ymax=1.02,
            xtick={70, 60, 50, 40, 30, 20},
            ytick={0.78, 0.82, 0.86, 0.90, 0.94, 0.98, 1.02},
            legend style={at={(0.98,0.5)},anchor=south east},
            ymajorgrids=true,
            grid style=dashed,
            x dir=reverse
        ]
        
        \addplot[
            color=red,
            mark=square,
            ]
            coordinates {
            (20,0.789)(30,0.827)(40,0.836)(50,0.845)(60,0.883)(70,0.895)
            };
            
        \addplot[
            color=blue,
            mark=*,
            ]
            coordinates {
            (20,0.992)(30,1.004)(40,0.991)(50,0.993)(60,1.00)(70, 0.998)
            };
        
        \addlegendentry{TabPFGen}
        \addlegendentry{Mean}
            
        \end{axis}
        \end{tikzpicture}
        % \caption{Mfeat Morphological}
    \end{subfigure}%
    % \hfill
    \hspace{0.06\textwidth} % Adjust the spacing here
    \begin{subfigure}[b]{0.18\textwidth}
        \centering
        \begin{tikzpicture}[scale=0.4]
        \begin{axis}[
            title={Vehicle},
            xlabel={Percentage of Data Missing [\%]},
            ylabel={RMSE},
            xmin=20, xmax=70,
            ymin=0.84, ymax=1.02,
            xtick={70, 60, 50, 40, 30, 20},
            ytick={0.84, 0.87, 0.90, 0.93, 0.96, 0.99, 1.02},
            legend style={at={(0.98,0.5)},anchor=south east},
            ymajorgrids=true,
            grid style=dashed,
            x dir=reverse
        ]
        
        \addplot[
            color=red,
            mark=square,
            ]
            coordinates {
            (70,0.981)(60,0.938)(50,0.915)(40,0.886)(30,0.859)(20,0.841)
            };
            
        \addplot[
            color=blue,
            mark=*,
            ]
            coordinates {
            (70,1.011)(60,0.990)(50,0.983)(40,0.976)(30,0.979)(20, 0.997)
            };
        
        \addlegendentry{TabPFGen}
        \addlegendentry{Mean}
            
        \end{axis}
        \end{tikzpicture}
        % \caption{Vehicle}
    \end{subfigure}%
    % \hfill
    \hspace{0.06\textwidth} % Adjust the spacing here
    \begin{subfigure}[b]{0.18\textwidth}
        \centering
        \begin{tikzpicture}[scale=0.4]
        \begin{axis}[
            title={Banknote Authentication},
            xlabel={Percentage of Data Missing [\%]},
            ylabel={RMSE},
            xmin=20, xmax=70,
            ymin=0.90, ymax=1.02,
            xtick={70, 60, 50, 40, 30, 20},
            ytick={0.90, 0.92, 0.94, 0.96, 0.98, 1.00, 1.02},
            legend style={at={(0.98,0.5)},anchor=south east},
            ymajorgrids=true,
            grid style=dashed,
            x dir=reverse
        ]
        
        \addplot[
            color=red,
            mark=square,
            ]
            coordinates {
            (70,0.992)(60,0.986)(50,0.992)(40,0.946)(30,0.951)(20,0.906)
            };
            
        \addplot[
            color=blue,
            mark=*,
            ]
            coordinates {
            (70,0.996)(60,0.994)(50,1.000)(40,1.007)(30,1.015)(20, 1.014)
            };
        
        \addlegendentry{TabPFGen}
        \addlegendentry{Mean}
            
        \end{axis}
        \end{tikzpicture}
        % \caption{Banknote Authentication}
    \end{subfigure}
    \caption{Imputation results. TabPFGen consistently has lower RMSE than the baseline method and also has decreasing RMSE with a decreasing fraction of missing data.}
    \label{fig: imputation}
\end{figure}
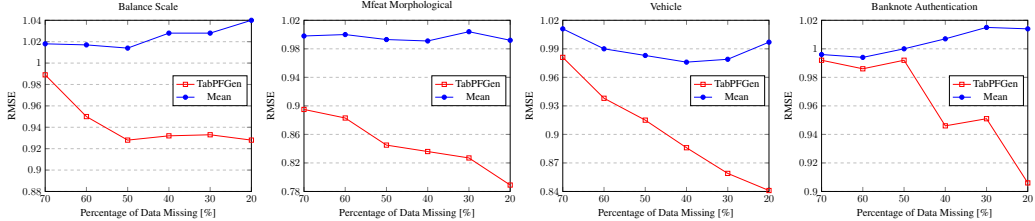

We conduct imputation analysis on 4 randomly selected datasets including: balance scale, mfeat morphological, vehicle, and banknote authentication. Details of the datasets can be found in Appendix \ref{section:dataset}. For all datasets, we uniform randomly mask out a certain percentage of the entries in the table as NaNs. For the baseline method, we simply use the mean of the entire column as the predicted value.

For our method, we initialize the missing entries as the mean value of the feature.
Then we start sampling with TabPFGen on $\xs$, but we fix the values of the non-missing features. Finally, we measure the RMSE between the original values and the sampled values for all missing entries. As shown in Figure \ref{fig: imputation}, for all datasets, TabPFGen results in lower RMSE than using the mean feature value as imputation. 
We can also observe that TabPFGen is able to reduce RMSE further as the percentage of data missing decreases, while the mean imputation method does not improve.
Even though these experiments do not compare to a robust set of baselines, they nevertheless demonstrate that using TabPFGen for imputation is a fruitful direction for further study.

\subsection{Ablation Studies}
\label{section: ablation}

\begin{table}[h]
\caption{Average AUC over OpenML-CC18 with error bars over 3 runs. Results shown across different TabPFGen variants. Top 4 rows: synthetic data as augmentation. Bottom 4 rows: synthetic data as replacement.}
\label{tab:ablation}
  \centering
  \scriptsize % Smaller font size for the table
  \setlength{\tabcolsep}{20pt} % Reduce the column padding by half
  \begin{tabular}{@{}lcccccccc@{}}
    \toprule
    Model & TabPFGen\textsubscript{core} & TabPFGen \\
    \midrule
    XGB 
    & $\mathbf{0.934} \scriptscriptstyle \pm 5e-4$ 
    & $\mathbf{0.934}  \scriptscriptstyle \pm 3e-4$ \\
    
    RF 
    & $\mathbf{0.913}  \scriptscriptstyle \pm 9e-4$ 
    & $0.912 \scriptscriptstyle \pm 4e-4$ \\
    
    LR 
    & $0.919 \scriptscriptstyle \pm 6e-4$ 
    & $\mathbf{0.921}  \scriptscriptstyle \pm 2e-4$ \\
    
    TabPFN 
    & $\mathbf{0.937} \scriptscriptstyle \pm 3e-4$ 
    & $0.935 \scriptscriptstyle \pm 3e-4$ \\

    \midrule
    
    XGB  
    & $\mathbf{0.927} \scriptscriptstyle \pm 1e-3$ 
    & $\mathbf{0.927} \scriptscriptstyle \pm 3e-4$ \\
    
    RF 
    & $\mathbf{0.906} \scriptscriptstyle \pm 7e-4$ 
    & $\mathbf{0.906} \scriptscriptstyle \pm 6e-4$ \\
    LR 
    & $0.918 \scriptscriptstyle \pm 6e-4$ 
    & $\mathbf{0.920} \scriptscriptstyle \pm 1e-3$ \\
    TabPFN 
    & $\mathbf{0.934} \scriptscriptstyle \pm 1e-3$ 
    & $\mathbf{0.934} \scriptscriptstyle \pm 2e-4$ \\
    \bottomrule
  \end{tabular}
\end{table}

% % \begin{table}[h]
% % \caption{Ablation Studies}
% % \label{tab:ablation}
% %   \centering
% %   \begin{tabular}{|c||c|c|c|c|}
% %     \hline
% %     \multirow{2}{*}{} &
% %       \multicolumn{4}{c|}{Synthetic as Replacement Ablation} \\
% %     & TabPFGen\textsubscript{core,SGLD} & TabPFGen\textsubscript{SGLD} & TabPFGen\textsubscript{core} & TabPFGen \\
% %     \hhline{|-|-|-|-|-|} 
% %     XGB & 0.859 & 0.825 & 0.905 & \textbf{0.913} \\
% %     \hline
% %     RF & 0.849 & 0.799 & 0.897 & \textbf{0.897} \\
% %     \hline
% %     LR & 0.835 & 0.863 & 0.914 & \textbf{0.916} \\
% %     \hline
% %   \end{tabular}
% % \end{table}

% % Ablation studies in Table \ref{tab:ablation} show that having an augmented energy and replacing SGLD with Adam + noise give gradual and steady improvement. 
We conduct ablation studies over the 2 TabPFGen variants in Table \ref{tab:ablation}, recalling that the Core variant is the basic technique outlined in Algorithm \ref{alg:tabpfgen_algorithm}, and the full TabPFGen variant also includes the swapping of $(\xs, \ys)$ and $(\xtr, \ytr)$ as described previous.
The top 4 rows indicate the performance of different TabPFGen variants when augmenting the training set with synthetic samples, and the bottom 4 rows show the downstream model performance when trained only on synthetic data. % The descriptions of different variants are included in Appendix \ref{section:tabpfgen_details}.
We see that the Core variant is generally competitive with the full TabPFGen -- even outperforming in some instances -- but we decided to use the full TabPFGen as the main approach because of its performance on the replacement task.

% While it can be observed that the TabPFGen variant described in the main text has the best performance overall, slightly outperforming the other variants, all four variants consistently outperform all other generative baselines (see Table~\ref{tab:augmentation_table}). With the exception of the LR model on the TabPFGen\textsubscript{core} variant, augmenting the training data with synthetic TabPFGen samples also consistently boosts performance over training on the original training set. These results demonstrates that the high performance of TabPFGen is not a result of specific parameter choices or sampling processes.

\subsection{Additional Class Balancing Experiment Details}
\label{section:additional_class_balancing}

\FloatBarrier

\begin{table}[h!]
  \caption{Average AUC using synthetic data for class balancing. Error bars shown over 3 runs.}
  \label{tab:class_balancing_std}
  \centering
  \scriptsize % Smaller font size for the table
  \setlength{\tabcolsep}{1.2pt} % Reduce the column padding by half
  \begin{tabular}{@{}lcccccccccc@{}}
    \toprule
     & Original & Sampling & \ttfamily{SMOTE} & \ttfamily{CTGAN} & \ttfamily{TVAE} & \ttfamily{NF} & \ttfamily{RTVAE} & \ttfamily{TabDDPM} & \ttfamily{TabPFGen} \\
    \midrule
    KC & $0.823 \scriptscriptstyle \pm 1e-3$ & $0.875 \scriptscriptstyle \pm 4e-4$ & $0.872 \scriptscriptstyle \pm 2e-4$ & $0.859 \scriptscriptstyle \pm 7e-4$ & $0.848 \scriptscriptstyle \pm 6e-4$ & $0.862 \scriptscriptstyle \pm 9e-4$ & $0.866 \scriptscriptstyle \pm 2e-3$ & $0.805 \scriptscriptstyle \pm 8e-4$ & $\mathbf{0.877} \scriptscriptstyle \pm 6e-4$ \\
    PC & $0.824 \scriptscriptstyle \pm 4e-3 $ & $0.811 \scriptscriptstyle \pm 4e-3$ & $0.836 \scriptscriptstyle \pm 4e-4$ & $0.827 \scriptscriptstyle \pm 4e-4$ & $0.835 \scriptscriptstyle \pm 5e-3$ & $0.825 \scriptscriptstyle \pm 1e-3$ & $\mathbf{0.841} \scriptscriptstyle \pm 5e-4$ & $0.825 \scriptscriptstyle \pm 5e-3$ & $\mathbf{0.841} \scriptscriptstyle \pm 8e-4$ \\
    BL & $0.731 \scriptscriptstyle \pm 4e-4$ & $0.757 \scriptscriptstyle \pm 4e-3$ & $0.756 \scriptscriptstyle \pm 3e-3$ & $0.743 \scriptscriptstyle \pm 7e-4$ & $0.714 \scriptscriptstyle \pm 4e-4$ & $0.755 \scriptscriptstyle \pm 5e-4$  & $0.706 \scriptscriptstyle \pm 4e-3$ & $\mathbf{0.774} \scriptscriptstyle \pm 4e-4$ & $0.767 \scriptscriptstyle \pm 4e-4$ \\
    CL & $0.925 \scriptscriptstyle \pm 5e-4$ & $0.935 \scriptscriptstyle \pm 3e-3$ & $0.949 \scriptscriptstyle \pm 1e-3$ & $0.793 \scriptscriptstyle \pm 4e-4$ & $0.771 \scriptscriptstyle \pm 4e-4$ & $0.795 \scriptscriptstyle \pm 4e-4$ & $0.909 \scriptscriptstyle \pm 4e-4$ & $0.915 \scriptscriptstyle \pm 5e-4$ & $\mathbf{0.955} \scriptscriptstyle \pm 3e-3$ \\
    DI & $0.837 \scriptscriptstyle \pm 4e-3$ & $0.832 \scriptscriptstyle \pm 4e-4$ & $0.832 \scriptscriptstyle \pm 7e-4$ & $0.831 \scriptscriptstyle \pm 5e-4$ & $0.813 \scriptscriptstyle \pm 5e-4$ & $0.832 \scriptscriptstyle \pm 3e-3$ & $0.837 \scriptscriptstyle \pm 5e-4$ & $0.843 \scriptscriptstyle \pm 4e-4$ & $\mathbf{0.844} \scriptscriptstyle \pm 4e-3$ \\
    \bottomrule
  \end{tabular}
\end{table}

\begin{table}[h!]
  \caption{Information of datasets used for class-balancing experiments}
  \label{tab:class_balancing_datasets}
  \centering
  \scriptsize % Smaller font size for the table
  \setlength{\tabcolsep}{6pt} % Adjust column padding
  
  \begin{tabular}{@{}lccc@{}}
    \toprule
    Dataset & Full Name & MinMaj \\
    \midrule
    KC & kc2 & 0.26 \\
    PC & pc3 & 0.11 \\
    BL & blood-transfusion-service-center & 0.31 \\
    CL & climate-model-simulation-crashes & 0.09 \\
    DI & diabetes & 0.54 \\
    \bottomrule
  \end{tabular}
\end{table}

Table \ref{tab:class_balancing_std} shows class balancing results with multiple runs for each experimental setup. The datasets KC, PC, BL, CL, and DI represent kc2, pc3, blood-transfusion-service-center, climate-model-simulation-crashes, and diabetes datasets, respectively. All of them are chosen from the OpenML-CC18 suite. We find that TabPFGen generally outperforms baseline methods, with only TabDDPM on blood-transfusion outperforming TabPFGen throughout the entire suite of experiments. 

Table \ref{tab:class_balancing_datasets} shows the full name for each dataset along with the ratio of minority class size to majority class size (MinMaj).

\subsection{Synthetic Data as Augmentation Details}
\label{section:details_synthetic_data_augmentation}

\begin{table}[h!]
  \caption{Detailed ``synthetic data as augmentation'' experiment using XGBoost}
  \label{tab:experiment_details_augment_xgb}
  \centering
  \scriptsize % Smaller font size for the table
  \setlength{\tabcolsep}{3pt} % Reduce the column padding by half
  \begin{tabular}{@{}lcccccccccc@{}}
    \toprule
    Dataset & Original & SMOTE & CTGAN & TVAE & NF & RTVAE & TabDDPM & TabPFGen \\
    \midrule
    balance-scale & $\mathbf{0.9939}$ & $0.9931$ & $0.8374$ & $0.8236$ & $0.8135$ & $0.8236$ & $0.9620$ & $0.9931$ \\
    mfeat-fourier & $\mathbf{0.9803}$ & $0.9801$ & $0.9761$ & $0.9773$ & $0.9754$ & $0.9770$ & $0.9768$ & $\mathbf{0.9803}$ \\
    mfeat-karhunen & $0.9983$ & $\mathbf{0.9989}$ & $0.9977$ & $0.9974$ & $0.9961$ & $0.9977$ & $0.9977$ & $0.9985$ \\
    mfeat-morphological & $0.9612$ & $0.9608$ & $0.9574$ & $0.9556$ & $0.9562$ & $0.9600$ & $0.9612$ & $\mathbf{0.9613}$ \\
    mfeat-zernike & $0.9735$ & $\mathbf{0.9749}$ & $0.9729$ & $0.9731$ & $0.9725$ & $0.9733$ & $0.9721$ & $0.9740$ \\
    diabetes & $0.8378$ & $0.8109$ & $0.8390$ & $0.8273$ & $0.8384$ & $\mathbf{0.8401}$ & $0.8295$ & $0.8378$ \\
    vehicle & $\mathbf{0.9282}$ & $\mathbf{0.9282}$ & $0.9219$ & $0.9248$ & $0.9147$ & $0.9074$ & $0.9234$ & $0.9272$ \\
    analcatdata\_authorship & $0.9997$ & $\mathbf{1.0000}$ & $0.9999$ & $0.9999$ & $0.9997$ & $0.9998$ & $0.9999$ & $0.9999$ \\
    pc4 & $0.9291$ & $0.9291$ & $0.9245$ & $0.9259$ & $0.9209$ & $0.9230$ & $0.9279$ & $\mathbf{0.9351}$ \\
    pc3 & $0.8288$ & $0.8261$ & $0.8187$ & $0.8251$ & $0.8304$ & $0.8323$ & $0.8279$ & $\mathbf{0.8392}$ \\
    kc2 & $0.8227$ & $0.8567$ & $0.8760$ & $0.8762$ & $0.8793$ & $\mathbf{0.8831}$ & $0.8801$ & $0.8716$ \\
    pc1 & $0.8489$ & $0.8683$ & $0.8428$ & $0.8465$ & $0.8602$ & $0.8556$ & $0.8800$ & $\mathbf{0.8971}$ \\
    banknote-authentication & $\mathbf{1.0000}$ & $0.9999$ & $0.9990$ & $0.9974$ & $0.9990$ & $0.9997$ & $\mathbf{1.0000}$ & $\mathbf{1.0000}$ \\
    blood-transfusion-service-center & $0.7312$ & $0.7495$ & $0.7221$ & $0.7662$ & $0.7626$ & $0.7580$ & $\mathbf{0.7651}$ & $0.7579$ \\
    qsar-biodeg & $0.9191$ & $0.9151$ & $0.9120$ & $0.9091$ & $0.9078$ & $0.9113$ & $0.9181$ & $\mathbf{0.9212}$ \\
    wdbc & $\mathbf{0.9904}$ & $0.9800$ & $0.9512$ & $0.9491$ & $0.9687$ & $0.9674$ & $0.9798$ & $0.9855$ \\
    steel-plates-fault & $0.9656$ & $0.9653$ & $0.9597$ & $\mathbf{0.9668}$ & $0.9627$ & $0.9604$ & $0.9585$ & $0.9654$ \\
    climate-model-simulation-crashes & $0.9255$ & $0.9447$ & $0.9081$ & $0.9106$ & $0.8586$ & $0.9389$ & $0.9347$ & $\mathbf{0.9586}$ \\
    \bottomrule
  \end{tabular}
\end{table}

\begin{table}[h!]
  \caption{Detailed ``synthetic data as augmentation experiment'' using Random Forest.}
  \label{tab:experiment_details_augment_rf}
  \centering
  \scriptsize % Smaller font size for the table
  \setlength{\tabcolsep}{3pt} % Reduce the column padding by half
  \begin{tabular}{@{}lcccccccccc@{}}
    \toprule
    Dataset & Original & SMOTE & CTGAN & TVAE & NF & RTVAE & TabDDPM & TabPFGen \\
    \midrule
    balance-scale & $0.8001$ & $0.8142$ & $0.7491$ & $0.7717$ & $0.7125$ & $0.7717$ & $0.7990$ & $\mathbf{0.8294}$ \\
    mfeat-fourier & $0.9794$ & $\mathbf{0.9806}$ & $0.9766$ & $0.9774$ & $0.9775$ & $0.9765$ & $0.9793$ & $0.9785$ \\
    mfeat-karhunen & $0.9968$ & $\mathbf{0.9974}$ & $0.9955$ & $0.9970$ & $0.9946$ & $0.9963$ & $0.9971$ & $0.9973$ \\
    mfeat-morphological & $0.9509$ & $0.9482$ & $0.9471$ & $0.9459$ & $0.9449$ & $0.9510$ & $\mathbf{0.9529}$ & $0.9502$ \\
    mfeat-zernike & $\mathbf{0.9714}$ & $0.9672$ & $0.9662$ & $0.9664$ & $0.9677$ & $0.9684$ & $0.9694$ & $0.9707$ \\
    diabetes & $0.8159$ & $0.8020$ & $0.8068$ & $0.7813$ & $0.8087$ & $0.8070$ & $0.8209$ & $\mathbf{0.8307}$ \\
    vehicle & $0.9188$ & $0.9153$ & $0.9155$ & $0.9183$ & $0.9109$ & $0.9167$ & $\mathbf{0.9237}$ & $0.9175$ \\
    analcatdata\_authorship & $0.9998$ & $\mathbf{0.9999}$ & $0.9998$ & $\mathbf{0.9999}$ & $0.9996$ & $\mathbf{0.9999}$ & $0.9995$ & $\mathbf{0.9999}$ \\
    pc4 & $0.9220$ & $0.9225$ & $0.9131$ & $0.9213$ & $0.9145$ & $0.9166$ & $\mathbf{0.9277}$ & $0.9261$ \\
    pc3 & $0.8047$ & $0.8173$ & $0.8101$ & $0.8090$ & $0.8085$ & $\mathbf{0.8318}$ & $0.7906$ & $0.8200$ \\
    kc2 & $0.8348$ & $\mathbf{0.8594}$ & $0.8259$ & $0.8497$ & $0.8484$ & $0.8230$ & $0.8377$ & $0.8348$ \\
    pc1 & $0.8853$ & $0.8851$ & $0.8740$ & $0.9063$ & $0.8902$ & $\mathbf{0.9066}$ & $0.8919$ & $0.8867$ \\
    banknote-authentication & $0.9996$ & $0.9998$ & $0.9992$ & $0.9992$ & $\mathbf{1.0000}$ & $0.9994$ & $\mathbf{1.0000}$ & $0.9998$ \\
    blood-transfusion-service-center & $0.7016$ & $0.6914$ & $0.6868$ & $0.7081$ & $0.6905$ & $0.7136$ & $\mathbf{0.7261}$ & $0.6984$ \\
    qsar-biodeg & $0.9158$ & $0.9156$ & $0.9076$ & $0.9106$ & $0.9076$ & $0.9072$ & $\mathbf{0.9198}$ & $0.9158$ \\
    wdbc & $0.9838$ & $0.9856$ & $0.9837$ & $0.9897$ & $0.9844$ & $\mathbf{0.9906}$ & $0.9860$ & $0.9802$ \\
    steel-plates-fault & $0.9577$ & $\mathbf{0.9601}$ & $0.9579$ & $0.9600$ & $0.9579$ & $0.9584$ & $0.9578$ & $0.9574$ \\
    climate-model-simulation-crashes & $0.8758$ & $0.8580$ & $0.8581$ & $0.8614$ & $0.7908$ & $0.8951$ & $\mathbf{0.9301}$ & $0.9261$ \\
    \bottomrule
  \end{tabular}
\end{table}

\begin{table}[h!]
  \caption{Detailed ``synthetic data as augmentation experiment'' using Logistic Regression.}
  \label{tab:experiment_details_augment_lr}
  \centering
  \scriptsize % Smaller font size for the table
  \setlength{\tabcolsep}{3pt} % Reduce the column padding by half
  \begin{tabular}{@{}lcccccccccc@{}}
    \toprule
    Dataset & Original & SMOTE & CTGAN & TVAE & NF & RTVAE & TabDDPM & TabPFGen \\
    \midrule
    balance-scale & $\mathbf{0.9574}$ & $0.9568$ & $0.8933$ & $0.8778$ & $0.8517$ & $0.8778$ & $0.9556$ & $0.9559$ \\
    mfeat-fourier & $0.9631$ & $0.9612$ & $0.9611$ & $\mathbf{0.9636}$ & $0.9605$ & $0.9601$ & $0.9570$ & $0.9584$ \\
    mfeat-karhunen & $0.9933$ & $0.9930$ & $0.9903$ & $0.9895$ & $0.9885$ & $0.9907$ & $0.9863$ & $\mathbf{0.9948}$ \\
    mfeat-morphological & $\mathbf{0.9664}$ & $0.9654$ & $0.9578$ & $0.9538$ & $0.9534$ & $0.9526$ & $0.9629$ & $0.9660$ \\
    mfeat-zernike & $\mathbf{0.9759}$ & $0.9750$ & $0.9754$ & $0.9757$ & $0.9708$ & $0.9744$ & $0.9645$ & $0.9757$ \\
    diabetes & $0.8384$ & $0.8292$ & $0.8330$ & $0.8304$ & $\mathbf{0.8440}$ & $0.8255$ & $0.8132$ & $0.8365$ \\
    vehicle & $0.9369$ & $\mathbf{0.9404}$ & $0.8914$ & $0.8917$ & $0.8607$ & $0.8803$ & $0.7469$ & $0.9388$ \\
    analcatdata\_authorship & $0.9999$ & $0.9999$ & $0.9832$ & $0.9909$ & $0.9976$ & $0.9782$ & $0.9989$ & $\mathbf{0.9999}$ \\
    pc4 & $0.8880$ & $0.8869$ & $\mathbf{0.8922}$ & $0.8743$ & $0.8597$ & $0.8615$ & $0.8500$ & $0.8830$ \\
    pc3 & $\mathbf{0.8201}$ & $0.8010$ & $0.7649$ & $0.7986$ & $0.8031$ & $0.8129$ & $0.7144$ & $0.8162$ \\
    kc2 & $0.8524$ & $0.8084$ & $0.8640$ & $0.8428$ & $\mathbf{0.8781}$ & $0.8497$ & $0.8283$ & $0.8727$ \\
    pc1 & $\mathbf{0.8233}$ & $0.8015$ & $0.7731$ & $0.8393$ & $0.8072$ & $0.8185$ & $0.7827$ & $0.8224$ \\
    banknote-authentication & $\mathbf{0.9999}$ & $\mathbf{0.9999}$ & $0.9981$ & $0.9983$ & $0.9986$ & $0.9918$ & $\mathbf{0.9999}$ & $\mathbf{0.9999}$ \\
    blood-transfusion-service-center & $0.7616$ & $0.7610$ & $0.7487$ & $\mathbf{0.7682}$ & $0.7669$ & $0.7520$ & $0.7610$ & $0.7639$ \\
    qsar-biodeg & $0.9060$ & $0.9046$ & $0.9037$ & $0.9055$ & $0.8789$ & $0.9024$ & $0.8664$ & $\mathbf{0.9073}$ \\
    wdbc & $0.9834$ & $0.9836$ & $0.9832$ & $0.9805$ & $0.9873$ & $\mathbf{0.9896}$ & $0.9729$ & $0.9841$ \\
    steel-plates-fault & $0.9375$ & $0.9374$ & $0.9189$ & $0.9295$ & $0.9175$ & $0.9247$ & $\mathbf{0.9500}$ & $0.9377$ \\
    climate-model-simulation-crashes & $0.9621$ & $0.9544$ & $0.9453$ & $0.9581$ & $0.9037$ & $\mathbf{0.9687}$ & $0.8192$ & $0.9610$ \\
    \bottomrule
  \end{tabular}
\end{table}

\begin{table}[h!]
  \caption{Detailed ``synthetic data as augmentation experiment'' using TabPFN.}
  \label{tab:experiment_details_augment_tabpfn}
  \centering
  \scriptsize % Smaller font size for the table
  \setlength{\tabcolsep}{3pt} % Reduce the column padding by half
  \begin{tabular}{@{}lcccccccccc@{}}
    \toprule
    Dataset & Original & SMOTE & CTGAN & TVAE & NF & RTVAE & TabDDPM & TabPFGen \\
    \midrule
    balance-scale & $\mathbf{0.9984}$ & $0.9983$ & $0.9510$ & $0.9582$ & $0.9508$ & $0.9582$ & $0.9792$ & $0.9955$ \\
    mfeat-fourier & $\mathbf{0.9810}$ & $0.9804$ & $0.9783$ & $0.9774$ & $0.9764$ & $0.9773$ & $0.9805$ & $0.9802$ \\
    mfeat-karhunen & $0.9980$ & $0.9979$ & $0.9976$ & $0.9978$ & $0.9975$ & $0.9976$ & $\mathbf{0.9986}$ & $0.9978$ \\
    mfeat-morphological & $\mathbf{0.9646}$ & $0.9637$ & $0.9644$ & $0.9627$ & $0.9638$ & $0.9630$ & $0.9638$ & $0.9638$ \\
    mfeat-zernike & $0.9818$ & $0.9817$ & $0.9806$ & $0.9807$ & $0.9816$ & $0.9810$ & $\mathbf{0.9819}$ & $\mathbf{0.9817}$ \\
    diabetes & $\textbf{}{0.8428}$ & $0.7766$ & $0.8423$ & $0.8307$ & $0.8418$ & $0.8308$ & $0.8224$ & $0.8138$ \\
    vehicle & $0.9568$ & $0.9516$ & $0.9419$ & $0.9477$ & $0.9467$ & $0.9450$ & $\mathbf{0.9541}$ & $0.9523$ \\
    analcatdata\_authorship & $\mathbf{1.0000}$ & $\mathbf{1.0000}$ & $\mathbf{1.0000}$ & $0.9999$ & $0.9999$ & $0.9999$ & $0.9999$ & $0.9999$ \\
    pc4 & $0.9233$ & $0.9283$ & $0.9320$ & $0.9326$ & $0.9254$ & $0.9306$ & $0.9146$ & $\mathbf{0.9329}$ \\
    pc3 & $0.8481$ & $0.8183$ & $0.8396$ & $0.8449$ & $0.8369$ & $0.8516$ & $0.8365$ & $\mathbf{0.8533}$ \\
    kc2 & $0.8777$ & $0.8565$ & $0.8816$ & $0.8701$ & $0.8779$ & $0.8615$ & $\mathbf{0.8831}$ & $0.8750$ \\
    pc1 & $\mathbf{0.8919}$ & $0.8782$ & $0.8709$ & $0.8864$ & $0.8753$ & $0.8888$ & $0.8767$ & $0.8903$ \\
    banknote-authentication & $\mathbf{1.0000}$ & $\mathbf{1.0000}$ & $0.9997$ & $\mathbf{1.0000}$ & $\mathbf{1.0000}$ & $\mathbf{1.0000}$ & $\mathbf{1.0000}$ & $\mathbf{1.0000}$ \\
    blood-transfusion-service-center & $0.7590$ & $0.7277$ & $0.7444$ & $0.7599$ & $0.7601$ & $0.7622$ & $\mathbf{0.7687}$ & $0.7423$ \\
    qsar-biodeg & $0.9306$ & $0.9242$ & $0.9291$ & $0.9220$ & $0.9242$ & $0.9214$ & $0.9287$ & $\mathbf{0.9293}$ \\
    wdbc & $0.9904$ & $0.9909$ & $0.9897$ & $0.9883$ & $0.9922$ & $0.9896$ & $\mathbf{0.9933}$ & $0.9892$ \\
    steel-plates-fault & $0.9634$ & $0.9623$ & $0.9596$ & $0.9613$ & $0.9623$ & $0.9631$ & $0.9627$ & $\textbf{}{0.9636}$ \\
    climate-model-simulation-crashes & $0.9625$ & $0.9563$ & $0.9354$ & $0.9343$ & $0.8915$ & $0.9546$ & $0.8946$ & $\mathbf{0.9649}$ \\
    \bottomrule
  \end{tabular}
\end{table}

\FloatBarrier

In Table \ref{tab:experiment_details_augment_xgb}, \ref{tab:experiment_details_augment_rf}, \ref{tab:experiment_details_augment_lr} and \ref{tab:experiment_details_augment_tabpfn}, we show the performance of TabPFGen and baseline methods on each of the 18 datasets individually.
We display the average values over three runs, but omit error bars to preserve space.
We find that TabPFGen outperforms the baseline methods on the majority of the datasets.
Also notably, SMOTE performs quite well even when compared against modern deep learning methods, which is consistent with the recent work from \citet{manousakas2023usefulness}.

% \clearpage
\subsection{Experiments with Different Hyperparameters of Downstream Models}
\label{section:downstream_model_hyperparameters}

\begin{table}[h!]
  \caption{Synthetic data as augmentation with various downstream model hyperparameters. Bolding is done here just to highlight the best result up to three significant figures -- only one run of each setting was conducted.}
  \label{tab:experiment_different_hypers}
  \centering
  \scriptsize % Smaller font size for the table
  \setlength{\tabcolsep}{3pt} % Reduce the column padding by half
  \begin{tabular}{@{}lcccccccc@{}}
    \toprule
    Model & Original & SMOTE & CTGAN & TVAE & NF & RTVAE & TabDDPM & TabPFGen \\
    \midrule
    XGB 0 & $0.903$ & $0.897$ & $0.901$ & $0.897$ & $0.891$ & $0.903$ & $0.909$ & $\textbf{0.914}$ \\
    RF 0 & $0.902$ & $0.895$ & $0.900$ & $0.895$ & $0.892$ & $0.900$ & $0.903$ & $\textbf{0.920}$ \\
    LR 0 & $0.888$ & $0.894$ & $0.884$ & $0.886$ & $0.884$ & $0.888$ & $0.871$ & $\textbf{0.905}$ \\
    \midrule
    XGB 1 & $0.906$ & $0.902$ & $0.903$ & $0.904$ & $0.896$ & $0.906$ & $0.911$ & $\textbf{0.915}$ \\
    RF 1 & $0.908$ & $0.902$ & $0.897$ & $0.897$ & $0.891$ & $0.904$ & $0.911$ & $\textbf{0.919}$ \\
    LR 1 & $\textbf{0.920}$ & $0.913$ & $0.901$ & $0.908$ & $0.905$ & $0.905$ & $0.885$ & $\textbf{0.920}$ \\
    \midrule
    XGB 2 & $0.915$ & $0.914$ & $0.906$ & $0.909$ & $0.909$ & $0.911$ & $0.914$ & $\textbf{0.928}$ \\
    RF 2 & $0.906$ & $0.905$ & $0.902$ & $0.898$ & $0.897$ & $0.905$ & $0.911$ & $\textbf{0.922}$ \\
    LR 2 & $0.868$ & $0.870$ & $0.863$ & $0.865$ & $0.866$ & $0.869$ & $0.852$ & $\textbf{0.885}$ \\
    \midrule
    XGB 3 & $0.894$ & $0.897$ & $0.864$ & $0.862$ & $0.853$ & $0.871$ & $0.881$ & $\textbf{0.912}$ \\
    RF 3 & $0.909$ & $0.909$ & $0.904$ & $0.903$ & $0.900$ & $0.908$ & $0.914$ & $\textbf{0.921}$ \\
    LR 3 & $0.897$ & $0.894$ & $0.866$ & $0.898$ & $0.882$ & $0.893$ & $0.871$ & $\textbf{0.922}$ \\
    \bottomrule
  \end{tabular}
\end{table}

\begin{table*}[hp]
\caption{Different downstream model hyperparameters for Table \ref{tab:experiment_different_hypers}.}
\label{tab:hyperparams_expt}
\centering
\scriptsize
\begin{tabular}{@{}l||p{0.25\linewidth}|p{0.25\linewidth}|p{0.25\linewidth}@{}}
\toprule
 & XGB  & RF & LR \\
\midrule
    \ttfamily{Set 0} &
        \begin{tabular}[t]{@{}l@{}}
            alpha: 1.274e-10, \\
            colsample\_bylevel: 0.960, \\
            colsample\_bytree: 0.785, \\
            gamma: 5.737e-07, \\
            lambda: 3.493e-14, \\
            learning\_rate: 1.235e-06, \\
            max\_depth: 8, \\
            min\_child\_weight: 3.737e-11, \\
            n\_estimators: 971, \\
            subsample: 0.766
        \end{tabular} & 
        \begin{tabular}[t]{@{}l@{}}
            max\_depth: 39, \\
            max\_features: sqrt, \\
            min\_samples\_split: 5, \\
            n\_estimators: 34
        \end{tabular} &
        \begin{tabular}[t]{@{}l@{}}
            C: 1.362e-03, \\
            fit\_intercept: True, \\
            max\_iter: 320, \\
            penalty: l2
        \end{tabular} \\
\midrule
    \ttfamily{Set 1} &
        \begin{tabular}[t]{@{}l@{}}
            alpha: 2.165e-16, \\
            colsample\_bylevel: 0.975, \\
            colsample\_bytree: 0.865,\\
            gamma: 2.892e-13, \\
            lambda: 9.203e-14,\\
            learning\_rate: 1.922e-06,\\
            max\_depth: 6, \\
            min\_child\_weight: 1.312e-16,\\
            n\_estimators: 956, \\
            subsample: 0.432
    \end{tabular} & 
    \begin{tabular}[t]{@{}l@{}}
        max\_depth: 43, \\
        max\_features: sqrt, \\
        min\_samples\_split: 5, \\
        n\_estimators: 40
    \end{tabular} &
    \begin{tabular}[t]{@{}l@{}}
        C: 0.2776,\\
        fit\_intercept: True,\\
        max\_iter: 152, \\
        penalty: none
    \end{tabular} \\
\midrule
\midrule
    \ttfamily{Set 2} &
        \begin{tabular}[t]{@{}l@{}}
            alpha: 9.415e-07,\\
            colsample\_bylevel: 0.311,\\
            colsample\_bytree: 0.433, \\
            gamma: 9.377e-11,\\
            lambda: 2.719e-09, \\
            learning\_rate: 0.0313,\\
            max\_depth: 3,\\
            min\_child\_weight: 2.436e-10,\\
            n\_estimators: 343, \\
            subsample: 0.673
        \end{tabular} & 
        \begin{tabular}[t]{@{}l@{}}
            max\_depth: 39,\\
            max\_features: sqrt,\\
            min\_samples\_split: 5,\\
            n\_estimators: 94
        \end{tabular} &
        \begin{tabular}[t]{@{}l@{}}
            C: 7.744e-05, \\
            fit\_intercept: True, \\
            max\_iter: 137, \\
            penalty': l2
        \end{tabular} \\
\midrule
    \ttfamily{Set 3} &
        \begin{tabular}[t]{@{}l@{}}
            alpha: 5.717e-16,\\
            colsample\_bylevel: 0.686, \\
            colsample\_bytree: 0.336,\\
            gamma: 1.149e-15,\\
            lambda: 0.293, \\
            learning\_rate: 0.574, \\
            max\_depth: 2,\\
            min\_child\_weight: 2.729e-10,\\
            n\_estimators: 445,\\
            subsample: 0.278
    \end{tabular} & 
    \begin{tabular}[t]{@{}l@{}}
        max\_depth: 11,\\
        max\_features: sqrt,\\
        min\_samples\_split: 5, \\
        n\_estimators: 119
    \end{tabular} &
    \begin{tabular}[t]{@{}l@{}}
        C: 0.0266,\\
        fit\_intercept: False, \\
        max\_iter: 180,\\
        penalty: none
    \end{tabular} \\

\bottomrule
\end{tabular}
\end{table*}

We also experiment with different hyperparameters of the downstream models to verify the robustness of our approach to the specific choice of downstream model, ensuring that the superior performance of TabPFGen over other generative models is not sensitive to such choices.
Table \ref{tab:experiment_different_hypers} shows the results of the ``synthetic data as augmentation'' experiment with 4 different sets of hyperparameters each for downstream XGBoost (XGB), random forest (RF), and logistic regression (LR) models.
We have found that TabPFGen has the best result in all of the hyperparameter settings, showing that the results presented in the main text are robust to such choices.
The parameters are randomly chosen using ParameterSampler from the \texttt{scikit-learn} library to avoid biasing the algorithm.
We list the different hyperparameters we have used in Table \ref{tab:hyperparams_expt}.

% \begin{table}[h]
%     \caption{}
%   \centering
%   \small
%   \begin{tabular}{|c||c||c|c|c|c|c|}
%     \hline
%     \multirow{2}{*}{} &
%       \multicolumn{1}{c||}{} &
%       \multicolumn{5}{c|}{\textbf{Synthetic Data as Augmentation}} \\
%     Dataset & Original & \ttfamily{GReaT} & \ttfamily{TabDDPM} & \ttfamily{CTGAN} & \ttfamily{SMOTE} & \ttfamily{TabPFGen} \\
%     \hhline{|-|-|-|-|-|-|-|} 
%     XGB 0 &  &  & &  &  &  \\
%     \hline
%     XGB 1 &  &  & &  &  &  \\
%     \hline
%     XGB 2 &  &  &  &  &  &  \\
%     \hline
%     XGB 3 &  & &  & &  &   \\
%     \hline
%     XGB 4 &  &  &  & &  &  \\
%     \hline
%     RF 0 &  &  & &  &  &  \\
%     \hline
%     RF 1 &  &  & &  &  &  \\
%     \hline
%     RF 2 &  &  &  &  &  &  \\
%     \hline
%     RF 3 &  & &  & &  &   \\
%     \hline
%     RF 4 &  &  &  & &  &  \\
%     \hline
%     LR 0 &  &  & &  &  &  \\
%     \hline
%     LR 1 &  &  & &  &  &  \\
%     \hline
%     LR 2 &  &  &  &  &  &  \\
%     \hline
%     LR 3 &  & &  & &  &   \\
%     \hline
%     LR 4 &  &  &  & &  &  \\
%     \hline
%   \end{tabular}
% \end{table}

\begin{table*}[hp]
\caption{Hyperparameters of baseline models.}
\label{tab:baseline_hyperparams}
\centering
\scriptsize
    \begin{tabular}{@{}l||p{0.6\linewidth}@{}}
    \toprule
    Model & Hyperparameters \\
    \midrule
    \ttfamily{TVAE} &
        \begin{tabular}[t]{@{}l@{}}
        n\_units\_embedding=500, \\
        lr=5e-4, \\
        weight\_decay=1e-5, \\
        batch\_size=1000, \\
        decoder\_n\_layers\_hidden=2, \\
        decoder\_n\_units\_hidden=256, \\
        decoder\_nonlin="leaky\_relu", \\
        decoder\_dropout=0.1, \\
        encoder\_n\_layers\_hidden=3, \\
        encoder\_n\_units\_hidden=256, \\
        encoder\_nonlin="leaky\_relu", \\
        encoder\_dropout=0.1, \\
        loss\_factor=1, \\
        data\_encoder\_max\_clusters=10, \\
        clipping\_value=1, \\
        sampling\_patience=500
        \end{tabular} \\
    \midrule
    \ttfamily{RTVAE} &
        \begin{tabular}[t]{@{}l@{}}
        n\_units\_embedding=500, \\
        lr=0.001, \\
        weight\_decay=1e-5, \\
        batch\_size=200, \\
        decoder\_n\_layers\_hidden=3, \\
        decoder\_n\_units\_hidden=500, \\
        decoder\_nonlin="leaky\_relu", \\
        decoder\_dropout=0, \\
        encoder\_n\_layers\_hidden=3, \\
        encoder\_n\_units\_hidden=500, \\
        encoder\_nonlin="leaky\_relu", \\
        encoder\_dropout=0.1, \\
        data\_encoder\_max\_clusters=10, \\
        robust\_divergence\_beta=2, \\
        clipping\_value=1, \\
        sampling\_patience=500
        \end{tabular} \\
    \midrule
    \ttfamily{CTGAN} &
        \begin{tabular}[t]{@{}l@{}}
        generator\_n\_layers\_hidden=2, \\
        generator\_n\_units\_hidden=256, \\
        generator\_nonlin="relu", \\
        generator\_dropout=0.1, \\
        generator\_opt\_betas=(0.9, 0.999), \\
        discriminator\_n\_layers\_hidden=2, \\
        discriminator\_n\_units\_hidden=256, \\
        discriminator\_nonlin="leaky\_relu", \\
        discriminator\_n\_iter=1, \\
        discriminator\_dropout=0.1, \\
        discriminator\_opt\_betas=(0.9, 0.999), \\
        lr=5e-4, \\
        weight\_decay=1e-3, \\
        batch\_size=1000, \\
        clipping\_value=1, \\
        lambda\_gradient\_penalty=10, \\
        encoder\_max\_clusters=10, \\
        sampling\_patience=500
        \end{tabular} \\
    \midrule
    \ttfamily{NF} &
        \begin{tabular}[t]{@{}l@{}}
        n\_layers\_hidden=2, \\
        n\_units\_hidden=256, \\
        batch\_size=1000, \\
        num\_transform\_blocks=1, \\
        dropout=0.1, \\
        batch\_norm=False, \\
        num\_bins=8, \\
        tail\_bound=3, \\
        lr=5e-4, \\
        apply\_unconditional\_transform=True, \\
        base\_distribution="standard\_normal", \\
        linear\_transform\_type="permutation", \\
        base\_transform\_type="rq-autoregressive", \\
        encoder\_max\_clusters=10, \\
        n\_iter\_min=100, \\
        sampling\_patience=500
        \end{tabular} \\
    \midrule
    \ttfamily{TabDDPM} &
        \begin{tabular}[t]{@{}l@{}}
        is\_classification=True, \\
        lr=0.002, \\
        weight\_decay=0.0001, \\
        batch\_size=1024, \\
        gaussian\_loss\_type='mse', \\
        scheduler='cosine', \\
        model\_type='mlp', \\
        dim\_embed=128, \\
        continuous\_encoder='quantile', \\
        sampling\_patience=500
        \end{tabular} \\
    \bottomrule
    \end{tabular}
\end{table*}

\end{document}